\documentclass{article} 
\PassOptionsToPackage{dvipsnames,table}{xcolor}
\usepackage[preprint]{colm2025_conference}
\usepackage{microtype}
\usepackage{hyperref}
\usepackage{url}
\usepackage{booktabs}

\usepackage{lineno}

\definecolor{darkblue}{rgb}{0, 0, 0.5}
\hypersetup{colorlinks=true, citecolor=darkblue, linkcolor=darkblue, urlcolor=darkblue}

\usepackage{latexsym}
\usepackage[T1]{fontenc}
\usepackage[utf8]{inputenc}
\usepackage[utf8]{inputenc}
\usepackage{fix-cm}
\usepackage{inconsolata}
\usepackage{graphicx}
\usepackage{amsfonts}
\usepackage{amsmath}
\usepackage{cleveref}
\usepackage{nicefrac}
\usepackage{xspace}
\usepackage{enumitem}
\usepackage{wrapfig}
\usepackage[most]{tcolorbox}
\usepackage{listings}
\usepackage{multirow}
\usepackage{multicol}
\usepackage{siunitx}
\usepackage{subcaption}
\usepackage{nicematrix}
\usepackage{makecell}

\definecolor{NavyBlue}{rgb}{0.0, 0.0, 0.5}
\definecolor{DarkerNavy}{rgb}{0.0, 0.05, 0.3}
\definecolor{DeepNavy}{rgb}{0.0, 0.0, 0.4}
\definecolor{DustyBlue}{rgb}{0.2, 0.3, 0.4}

\tcbset{
  promptbox/.style={
    top=5pt,
    bottom=5pt,
    colback=lightgray!20,
    colframe=black,
    colbacktitle=DustyBlue,
    enhanced,
    center,
    breakable,
    attach boxed title to top center={yshift=-0.1in,xshift=0.0in},
    boxed title style={boxrule=0pt,colframe=white,},
    before upper={\parindent0pt\setlength{\parskip}{2pt}},
  }
}
\newtcolorbox{promptbox}[2][]{promptbox, title=#2,#1}

\hypersetup{
    colorlinks=true,
    linkcolor=darkblue,  
    citecolor=darkblue,  
    filecolor=magenta,      
    urlcolor=magenta,   
}

\title{CompassVerifier: A Unified and Robust Verifier for LLMs Evaluation and Outcome Reward}
\author{%
Shudong Liu$^{1,2,}$\thanks{\begin{minipage}[t]{\textwidth}
Equal contribution.
$^{\aleph}$Project lead.
$^{\dagger}$Corresponding authors.
Work done during Shudong's internship at Shanghai AI Laboratory.\\
Email: \texttt{nlp2ct.shudong@gmail.com; \{liuhongwei,zhangsongyang\}@pjlab.org.cn}
\end{minipage}}\;,
Hongwei Liu$^{1,*}$,
Junnan Liu$^{1}$,
Linchen Xiao$^{1}$,
\\
\textbf{Songyang Gao$^{1}$,
Chengqi Lyu$^{1}$,
Yuzhe Gu$^{1}$,
Wenwei Zhang$^{1}$,
}
\\
\textbf{Derek F. Wong$^{2,\dagger}$,
Songyang Zhang$^{1,\dagger,\aleph}$,
Kai Chen$^{1,\dagger}$
}
\\
\textsuperscript{$^{1}$}Shanghai AI Laboratory \quad \textsuperscript{$^{2}$}NLP$^{2}$CT Lab, University of Macau \\
}

\begin{document}

\ifcolmsubmission
\linenumbers
\fi

\maketitle

\begin{abstract}
Answer verification is crucial not only for evaluating large language models (LLMs) by matching their unstructured outputs against standard answers, but also serves as the reward model to guide LLM optimization. 
Most evaluation frameworks rely on regularized matching or employ general LLMs for answer verification, which demands extensive, repetitive customization for regex rules or evaluation prompts.
Two fundamental limitations persist in current methodologies: 1) the absence of comprehensive benchmarks that systematically evaluate verification capabilities across different LLMs; and 2) the nascent stage of verifier development, where existing approaches lack both the robustness to handle complex edge cases and the generalizability across different domains. 
In this work, we develop \textbf{CompassVerifier}, an accurate and robust lightweight verifier model for evaluation and outcome reward. It demonstrates multi-domain competency spanning math, knowledge, and diverse reasoning tasks, with the capability to process various answer types, including multi-subproblems, formulas, and sequence answers, while effectively identifying abnormal/invalid responses. We introduce \textbf{VerifierBench} benchmark comprising model outputs collected from multiple data sources, augmented through manual analysis of meta error patterns to enhance CompassVerifier. We anticipate that CompassVerifier and VerifierBench will facilitate answer verification, evaluation protocols, and reinforcement learning research. Code and dataset are available at \url{https://github.com/open-compass/CompassVerifier}.
\end{abstract}

\section{Introduction}
Answer verification plays a critical role in the evaluation and training of large language models (LLMs), particularly for objective questions with verifiable answers~\citep{achiam2023gpt,yang2024qwen2,liu2024deepseek,abs-2412-13147}. At the evaluation level, it enables precise measurement of performance differences across models~\citep{chang2024survey}; at the training level, it serves as a quality check for self-improvement~\citep{hosseini2024v,song2025mind}. With the rapid development of large reasoning models (LRMs) and reinforcement learning (RL), answer verification has further become a key component in constructing rule-based rewards, providing feedback signals to guide model optimization and iteration~\citep{guo2025deepseek,O1-Preview,luong2024reft,wang2025harnessing,zhong2025comprehensive}.

Existing answer verification methods can be broadly categorized into two types. The first type relies on regularized string matching, such as extracting content following \texttt{``The answer is”} to compare with reference answers, or using tools like math-verify~\citep{math-verify} to check formula equivalence in mathematical tasks. The second type employs general LLMs for consistency judgment, where a specific prompt is designed to instruct the model to evaluate the alignment between \textit{candidate} and \textit{reference answers}. However, both approaches suffer from significant limitations: the former requires repetitive customization of matching rules for different tasks and is prone to verification failures due to extraction errors; the latter demands frequent prompt adjustments to accommodate diverse tasks, domains, and answer types, while also facing the risk of misjudgment caused by model hallucination. Meanwhile, there is still no challenging benchmark available to evaluate and distinguish the verification capabilities of different models, nor to guide the development and iteration of verifiers.

In this paper, we establish a systematic framework for evaluating and training answer verification systems. We first introduce \textbf{VerifierBench}, a challenging benchmark for answer verification that aggregates numerous samples where rule-based methods frequently err or LLMs tend to produce incorrect judgments or hallucinations.
We integrated over one million data samples through the OpenCompass~\citep{2023opencompass} evaluation framework, encompassing responses from more than 50 models across 15 carefully selected datasets. Following large-scale data collection, each sample underwent a multi-stage filtering pipeline culminating in rigorous domain expert review and calibration.
VerifierBench facilitates precise measurement of verification capabilities across diverse models, addressing complex scenarios where both rule-based matching and general models often fail, and offering manually analyzed summaries of prevalent error patterns.

We further present \textbf{CompassVerifier}, a series of lightweight yet robust and accurate verification models. The training data originates from three key sources: 1) The original training set from VerifierBench, which undergoes multi-model validation with simple, easily verifiable samples removed; 2) Formula-enhanced data, where we leverage the powerful DeepSeek-V3 model to generate numerous equivalent complex formulas with corresponding reasoning processes to improve formulaic answer evaluation; 3) Hallucination-specific data, where we systematically analyze failure patterns from human validation cases and synthesize targeted training samples to address common hallucination errors.

\begin{figure*}[t]
    \centering
    \includegraphics[scale=0.53]{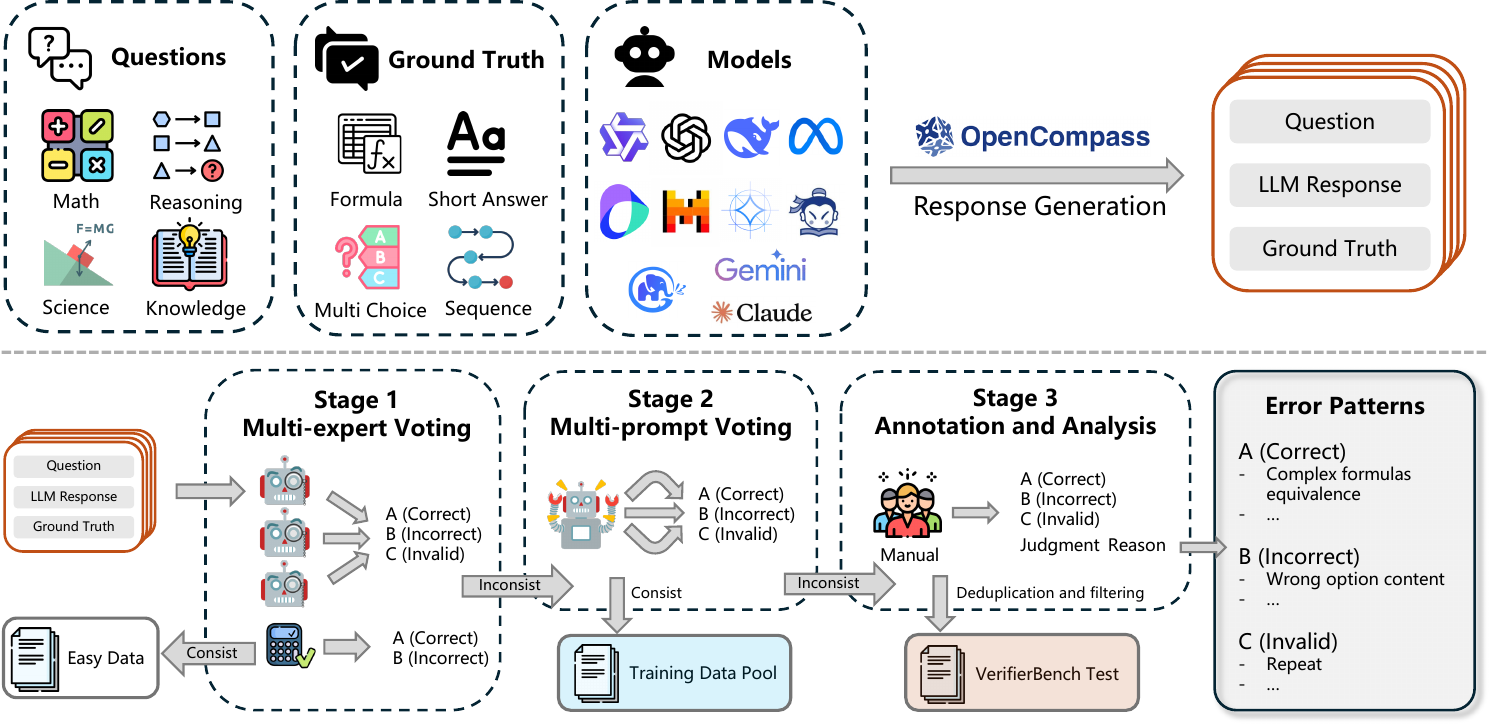}
    \caption{Overview of VerifierBench pipeline. Using OpenCompass~\citep{2023opencompass}, we collected more than 1 million LLM responses, applying multi-stage, multi-model verification with tool-assisted cleaning and filtering to create VerifierBench's test/base training sets and catalog common verification error patterns.}
    \label{fig_pipeline}
\end{figure*}

Our contributions are threefold:
\begin{itemize}[leftmargin=0.8em, topsep=0.7pt]
    \item We propose \textbf{VerifierBench}, a novel and challenging benchmark meticulously designed for fine-grained evaluation of verification abilities.
    \item We develop \textbf{CompassVerifier}, a series of robust and efficient verification models enhanced through our three proposed techniques, achieving state-of-the-art performance across diverse domains and tasks. CompassVerifier can also effectively serve as a reward model in RL training, delivering more precise and reliable feedback signals for policy optimization.
    \item Through a systematic analysis of prevalent failure modes in LLM-based verification, including characteristic hallucination phenomena and error propagation, we derive actionable insights aimed at advancing the design and robustness of future verification systems.
\end{itemize}


\section{Related Work}
\subsection{Answer Verification} 

Unlike traditional discriminative models with well-defined classification labels, the unstructured outputs of generative LLMs pose unique verification challenges~\citep{cobbe2021training}. Current approaches to verifying LLM-generated answers can be broadly categorized into \emph{outcome verification} and \emph{process verification}~\citep{kawabata2024rationale,zhang2025lessons}.

\emph{Outcome verification} focuses on assessing the correctness of final answers, typically through string-based pattern matching~\citep{2023opencompass,eval-harness,openai-evals}. Common practice instructs LLMs to output answers in predefined formats for character-level comparison with ground truth. For formulaic answers, specialized tools like Math-Verify~\citep{math-verify} have been developed to handle equivalence checking. However, due to the inherent unpredictability of LLM outputs, such methods often suffer from matching failures or inaccuracies. Many studies thus employ general LLMs as verifiers via tailored prompts. While effective, both methods demand task-specific customization through either regex patterns or verified prompts, creating labor-intensive workflows.
\emph{Process verification}, requiring detection of reasoning errors in intermediate steps, has seen recent advances in both LLM-based verifiers and evaluation benchmarks~\citep{lu2024autopsv, skyworkopeno12024, lightman2023lets, zheng2024processbench, zhou2024your}. However, process verifiers remain less frequently adopted in evaluations due to instability and high resource costs, and have not demonstrated substantially superior performance compared to outcome verifiers in RL.

We focus on scalable and robust outcome verification by developing a unified verifier that serves dual purposes: 1) as an evaluation model for benchmarking model performance, and 2) as a real-time reward model for RL training. By addressing the limitations of existing methods, such as ad-hoc prompt engineering and brittleness to output variations, CompassVerifier prioritizes efficiency, generalizability, and reliability across diverse tasks.

\subsection{LLM-as-a-Judge} 
The comprehensive capabilities of LLMs enable them to serve as cost-effective alternatives to human experts in evaluation tasks, a concept known as ``LLM-as-a-Judge”~\citep{gu2024survey,li2024generation}, which can be categorized into two approaches: \emph{subjective judgment} and \emph{objective judgment}.

\emph{Subjective judgment} typically operates in scenarios without ground-truth answers, where LLMs score individual responses (Pointwise)~\citep{zhu2025judgelm} or express preferences between paired responses (Pairwise)~\citep{wang2024pandalm}. This requires the LLM to evaluate various aspects of responses, including usefulness, harmlessness, and creativity, and even identify reasoning stepwise errors in the responses~\citep{cao2024compassjudger,li2024generative,alpaca_eval}. Recent studies also employ RL and inference-time scaling like generative critiques, long-CoT, and multi-sampling voting for judgment, albeit with high computational costs~\citep{liu2025inference,shi2025heimdall}. \emph{objective judgment} is a more straightforward approach, requiring only the evaluation of response correctness against ground-truth. Beyond simple string matching, the prevalent method employs large-scale LLMs with carefully designed evaluation prompts for judgment. Recently, to enable smaller models to achieve comparable verification capabilities to large LLMs, \citet{chen2025xverify} proposes xVerify and its accompanying benchmark, which trains smaller verifier models by distilling GPT-4o's capabilities. Other concurrent studies have also focused on distilling verification capabilities from large models to smaller ones to achieve better cost-effectiveness~\citep{ma2025generalreasoner,su2025expanding}.


We claim that objective judgment with ground-truth has yet to reach maturity, lacking both challenging benchmarks to discriminate model abilities and robust unified models. To address these gaps, we are committed to developing VerifierBench to rigorously test different models' verification capabilities and CompassVerifier to provide the research community with an accurate evaluation tool.

\section{VerifierBench}
The primary challenge in verifier development lies in the lack of comprehensive benchmarks and rigorous evaluation methodologies. Large-scale commercial models are often preferred for answer-matching tasks due to the prevailing assumption of scaling laws. However, critical questions remain unanswered: 1) To what extent do answer matching and objective judgment tasks adhere to scaling laws? 2) How should we balance model performance against computational costs in verification?

To answer these questions, in this work, we present VerifierBench, a systematic benchmark for evaluating diverse models' judgment and verification capabilities. VerifierBench addresses this gap through: 1) Large-scale data collection for answer matching (\ref{Data Collection}); 2) Multi-round validation involving multiple LLMs and human annotators (\ref{Data filtering pipeline}); 3) Case analysis of typical error patterns to identify failure modes (\ref{Statistics and Analysis}).

\subsection{Data Collection}
\label{Data Collection}
The crux of the answer verification task hinges on its capacity to encompass a comprehensive range of verifiable answer types and heterogeneous model responses.
To comprehensively gather such data, we employed the OpenCompass framework~\citep{2023opencompass} to conduct large-scale evaluations across multiple models and datasets. Our systematic approach yielded more than 1,325,293 samples covering four key domains: knowledge, mathematics, science, and general reasoning. The collected data features:
\begin{itemize}[leftmargin=1em, topsep=0pt]
 \item \textbf{Answer Type Diversity}: Multiple response formats including multiple-choice question options, mathematical formulations, short texts, multi-subproblem items, and long-sequence responses, etc.
 \item \textbf{Prompt Variability}: Input prompts covering few-shot, zero-shot, and dataset-specific formatting requirements.
 \item \textbf{Response Characteristics}: Model outputs ranging from short and long chain-of-thought (CoT) answers to direct responses and anomalous outputs (e.g., repetitions, truncations).
 \item \textbf{Diverse Model Coverage}: Comprehensive representation across commercial LLMs, open-source LLMs, and emerging LRMs, spanning diverse model scales.
\end{itemize}

Formally, our collected data consists of triplets: $\mathcal{D} = \{(q_i, a_i^*, r_i^m)\}_{i=1}^N$,
where $q_i \in \mathcal{Q}$ represents the $i$-th question, $a_i^* \in \mathcal{A}$ denotes the corresponding reference answer, $r_i^m \in \mathcal{R}$ is the response generated by model $m \in \mathcal{M}$. The primary objective of VerifierBench construction is to augment these triplets with verification labels, resulting in verified quadruples:

\begin{equation}
\mathcal{D}_{\text{VerifierBench}} = \{(q_i, a_i^*, r_i^m, v_i)\}_{i=1}^N,
\end{equation}

where $v_i \in \{\text{Correct}, \text{Incorrect}, \text{Invalid}\}$ is the verification label indicating the correctness of $r_i^m$ with respect to $a_i^*$. Notably, during data collection and curation, we identified numerous responses exhibiting abnormal or exceptional behaviors. These include abruptly truncated outputs, excessive repetition, and cases where models refused to answer due to ethical considerations or other constraints. We therefore categorize such instances as \emph{invalid} responses to enable a more fine-grained evaluation.

\subsection{Data Construction Pipeline}
\label{Data filtering pipeline}
Our multi-stage verification pipeline, integrating LLMs, human annotators, and rule-based tools, efficiently identifies high-value training and testing samples from a large collected dataset.

\vspace{-0.2em}
\paragraph{Multi-Expert Voting.} Initially, samples undergo direct verification (no CoT reasoning) by Qwen2.5-Instruct models (7B, 14B, 32B). Samples with consensus are deemed trivial cases reliably handled by weaker models and are removed, offering minimal value. For mathematical domains (Math, GSM8K, and AIME datasets), we also incorporated Math-Verify~\citep{math-verify} as an additional expert verifier.

\vspace{-0.5em}
\paragraph{Multi-prompt Voting.} Disputed samples advance to a second verification stage, where DeepSeek-V3 is employed with multiple prompts to generate diverse CoT reasoning paths. Consensus samples from this stage, representing moderately challenging instances, constitute our training pool. Our experiments revealed significant challenges in developing a universal verification prompt applicable across all datasets, evidenced by substantial residual disagreements after the second verification round. To address this, we implemented an additional verification phase for selected datasets, featuring domain-optimized prompts. For instance, the Chinese SimpleQA dataset required specially crafted Chinese-language prompts to achieve reliable verification outcomes.

\vspace{-0.5em}
\paragraph{Human Annotation and Analysis.} The remaining disputed samples are human-annotated, with high-value cases primarily allocated to the test set. For the VerifierBench test set, we systematically excluded proof-based questions, open-ended problems, and numerical answers with ambiguous acceptability thresholds. These non-binary judgment cases, requiring specialized verification tools or domain expertise, are deferred to future work, ensuring VerifierBench focuses on clearly verifiable samples. Finally, we get the VerifierBench dataset, and we also make sure there is no overlap between VerifierBench test set and the train set for training CompassVerifier model.

\vspace{-0.5em}
\paragraph{Identification of Flawed Samples.}
Human annotation also identified a distinct category: ``flawed samples”. Errors in these samples stem not from model deficiencies in problem-solving but from issues inherent to the questions (e.g., ambiguity, incorrect standard answers) or external factors (e.g., improper output truncation, generation of meaningless repetitive text, model refusal to answer).
Such flawed samples, if not distinguished, can skew model capability assessment and hinder effective model iteration. These issues are often overlooked in traditional evaluation paradigms.
Consequently, we explicitly label these samples as ``Invalid” and integrate them into the VerifierBench test set. This approach enables a more granular, multi-dimensional, and realistic perspective for model performance verification.


\subsection{Statistics and Analysis}
\label{Statistics and Analysis}


\noindent \textbf{Statistics.} We present the statistical characteristics of the VerifierBench test set across three dimensions: label categories (Table \ref{tab:VB_class}), problem domains (Table \ref{tab:VB_domain}), and answer types (Table \ref{tab:VB_answer_type}). After filtering and balancing, the dataset composition shows an approximate 4:6 ratio between Category A and B samples, with Category C representing about 7\% of the total. Regarding problem domains, general reasoning, and mathematical reasoning constitute the majority, aligning with the current needs of RL training on LLMs. Classified by DeepSeek-V3, the answer types comprise seven categories: multiple-choice, numerical values, short answers, formulas, multi-subproblem, sequences, and binary answers. The detailed dataset sources are provided in Table \ref{tab:VB_dataset}, with concrete examples illustrated in Section \ref{Cases_VerifierBench}. 

\noindent \textbf{Error Analysis and Patterns.} VerifierBench is designed not merely as a benchmark dataset for model evaluation, but as a comprehensive framework incorporating extensive \emph{human analysis} and \emph{case studies}.  During annotation, we required annotators to provide detailed judgment rationales in addition to final labels. Through systematic collection and analysis of these rationales, we identified and categorized over 30 meta error patterns (\Cref{Details of Meta Error Patterns}), which represent fundamental causes of mistakes and hallucinations in LLM-based answer verification. For example, while mathematically equivalent formulas are conventionally accepted as correct answers by LLMs or tools, they should be rejected for expression simplification problems. Similarly, for questions admitting multiple valid answers listed in the reference answer, a model response matching any one option should be considered correct, rather than complete coverage. We have found these meta patterns invaluable for both diagnostic analysis and targeted model improvement, and have incorporated them into our training framework. 

\begin{figure*}[t!]
    \centering
    \includegraphics[scale=0.56]{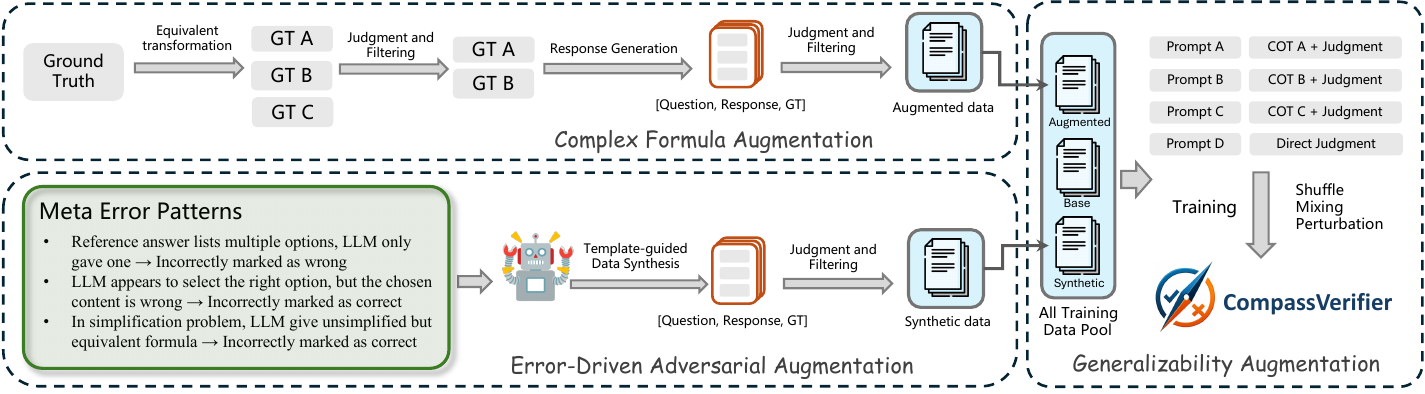}
    \caption{Overview of CompassVerifier training pipeline.}
    \label{fig_compasverifier}
\end{figure*}

\section{CompassVerifier}
CompassVerifier is designed to deliver efficient, high-performance, and robust answer verification. The system leverages filtered (question, reference answer, model response) triples from VerifierBench with golden judgments as training supervision. We also propose three key techniques to drive its performance: \textit{Complex Formula Augmentation} enhances formula variants verification, \textit{Error-Driven Adversarial Augmentation} fortifies against failures, and \textit{Generalizability Augmentation} ensures cross-domain and cross-prompt applicability. \Cref{fig_compasverifier} shows the whole pipeline of training CompassVerifier. Details of the composition of the training Data in \Cref{Details of CompassVerifier Model Train Data}.

\subsection{Error-Driven Adversarial Augmentation}
To address potential annotation inaccuracies in our filtered data (see \Cref{Data filtering pipeline}), we employ a three-phase adversarial augmentation strategy. 
\paragraph{Human-in-the-Loop Analysis.} Domain experts manually verify 5,000 annotated samples, identify and document failure rationales such as LLM misunderstandings of task constraints, misinterpretation of critical information in questions, and divergent penalty thresholds among judge models. 
\paragraph{Pattern Clustering.} We apply density-based clustering to these rationales, revealing over 20 high-impact error categories, particularly vulnerabilities in perspective-taking and format adherence. Analysis and details are shown in \Cref{Details of Meta Error Patterns}. 
\paragraph{Meta-Judge Template Generation.} For each error cluster, we develop structured templates that encode: \emph{1) Question Characteristics} (domain-specific requirements, content/format constraints) and \emph{2) Response Error Patterns} (failure types, localization, severity).

This aligns model judgments with human values and improves robustness against: (1) overstrict format-based rejection, (2) underpenalization of conceptual errors in fluent responses, and (3) context-sensitive scoring variations.

\subsection{Complex Formula Augmentation}

Verifying answers in domains such as the natural sciences is challenging due to the prevalence of complex expressions. These expressions often exhibit diverse notational conventions (e.g., symbolic, algebraic, floating-point, integer). Consequently, automated verifiers lacking robust mathematical equivalence understanding may erroneously reject semantically correct responses that differ superficially from reference solutions.

To address this issue, we introduce a \emph{Complex Formula Augmentation} strategy that systematically generates multiple, notation-variant answers for each problem instance. Our procedure is as follows:

\vspace{-0.5em}
\paragraph{Reference Normalization.} For each original question–answer pair in our dataset, we first convert the reference answer into a canonical representation, normalizing numeric precision and symbolic structure.

\vspace{-0.5em}
\paragraph{Variant Generation.} We leverage the DeepSeek-v3~\citep{ma2025generalreasoner} to produce between one and three alternative formulations of the canonical answer. These variants include: \emph{1) Symbolic rearrangements} (e.g., rationalizing denominators, applying algebraic identities). \emph{2) Precision-preserving floating-point expansions}. \emph{3) Equivalent integer or fraction representations}. We enforce strict constraints to avoid precision loss and ensure each variant remains mathematically equivalent to the original answer within the problem context.

\vspace{-0.5em}
\paragraph{Quality Control.} All generated variants are automatically checked for equivalence using a symbolic algebra engine, and a subset is manually reviewed by subject-matter experts to confirm correctness and naturalness of presentation.

By exposing the verifier to diverse but equivalent formulae, we markedly improve its ability to recognize correct answers regardless of notational differences, thereby reducing false negative rates in formula-intensive tasks.

\subsection{Generalizability Augmentation}

Existing verifier models often rely on task-specific prompts, limiting their generalizability across different problems and subtle answer variations (e.g., numerical precision in TheoremQA~\citep{chen-etal-2023-theoremqa}). To address this, we propose a \textit{Generalizability Augmentation} strategy to enhance adaptability by systematically expanding prompt and response diversity in training data. We collect diverse prompts from public datasets (e.g., TheoremQA, GPQA~\citep{rein2024gpqa}, GAOKAOBench~\citep{zhang2023evaluating}) and real-world scenarios, covering over 20 task types. For each prompt type, we design multiple variants, varying questioning styles, context lengths, linguistic registers, and instruction granularity. Our augmentation employs two key techniques:

\vspace{-0.5em}
\paragraph{Prompt Rewriting and Perturbation.} We use LLMs (e.g., DeepSeek-v3) to automatically generate paraphrases, structural modifications, and detail-enriched prompt variants, while maintaining consistency with the final judgment.
Furthermore, during training, we introduce prompt random sampling, dynamic mixing, and a prompt-invariance mechanism to prevent overfitting and encourage consistent judgments across different prompt formulations, thereby enhancing generalization.

\vspace{-0.5em}
\paragraph{Long-context Generalization.} To improve robustness in long-context scenarios, we apply various perturbations to responses collected from LRMs (e.g., DeepSeek-R1 and its distilled variants) in the training set, including truncating different portions (e.g., first 20\%, 40\%, 60\%) of the thinking process, replacing thinking tags (e.g., \texttt{<think>} or \texttt{</think>}) with alternative labels, or removing them entirely, while ensuring the final judgment remained consistent with the original response.

\section{Experiments}
\paragraph{Baselines and Setup.} We conduct comprehensive evaluations on VerifierBench across various model scales of CompassVerifier, ranging from 3B to 32B parameters. Baseline models include: (1) general LLMs such as Qwen2.5~\citep{yang2024qwen2}, Qwen3~\citep{yang2024qwen2}, DeepSeek-V3~\citep{guo2025deepseek}, and GPT-4o~\citep{GPT-4o}; and (2) two recently proposed specialized verifier models: xVerify~\citep{chen2025xverify} and Tencent-Qwen2.5-7B-Instruct-RLVR~\citep{su2025expanding}. We ask the model directly generate the final judgment of the given response and report F1 and Accuracy as metrics. More evaluation and training details are shown in \Cref{Details of CompassVerifier Experiments}.

\subsection{Main Results}
\begin{table*}[t]
    \centering
    \caption{Main results on the VerifierBench benchmark. For fair comparison, we treat the ``\textbf{Invalid}” instances in VerifierBench as incorrect labels, presenting results in a binary classification framework. We report Accuracy and F1 scores (\%) across four categories and their average.}
    \label{tab:main-results}
    \resizebox{0.99\textwidth}{!}{
    \begin{NiceTabular}{lcccccccc|cc}
    \toprule
    \multirow{2}{*}{Model} &
    \multicolumn{2}{c}{Math} &
    \multicolumn{2}{c}{General Reasoning} &
    \multicolumn{2}{c}{Knowledge} &
    \multicolumn{2}{c}{Science} &
    \multicolumn{2}{c}{Average} \\
    \cmidrule(r){2-3} \cmidrule(r){4-5} \cmidrule(r){6-7} \cmidrule(r){8-9} \cmidrule(l){10-11}
     & Acc. & F1 & Acc. & F1 & Acc. & F1 & Acc. & F1 & Acc. & F1 \\
    \midrule
    \rowcolor[gray]{0.88}
    \Block{1-11}{\textit{General LLMs}} & & & & & & & & & & \\
    \midrule
    Qwen2.5-7B-Instruct & 53.0 & 30.0 & 58.9 & 51.1 & 55.8 & 50.7 & 64.0 & 36.6 & 57.9 & 42.1 \\
    Qwen2.5-14B-Instruct & 51.6 & 37.4 & 57.3 & 44.9 & 50.9 & 37.8 & 70.0 & 47.9 & 57.4 & 42.0 \\
    Qwen2.5-32B-Instruct & 53.1 & 31.6 & 64.6 & 42.2 & 60.0 & 46.4 & 77.4 & 48.8 & 63.8 & 42.2 \\
    Qwen2.5-72B-Instruct & 57.0 & 37.5 & 61.4 & 49.0 & 70.0 & 68.5 & 77.9 & 60.5 & 66.6 & 53.9 \\
    \midrule
    Qwen3-8B & 53.0 & 51.6 & 61.6 & 61.8 & 63.8 & 69.4 & 57.9 & 42.9 & 59.1 & 56.4 \\
    Qwen3-14B & 65.1 & 44.1 & 76.8 & 66.7 & 69.8 & 66.7 & 81.6 & 56.8 & 73.3 & 58.6 \\
    Qwen3-30B-A3B & 59.7 & 62.4 & 63.4 & 63.2 & 61.5 & 64.4 & 59.5 & 48.7 & 61.0 & 59.7 \\
    Qwen3-32B & 64.4 & 54.6 & 74.9 & 70.3 & 68.7 & 69.5 & 74.7 & 52.8 & 70.7 & 61.8 \\
    Qwen3-235B-A22B & 64.2 & 53.9 & 78.5 & 73.7 & 67.4 & 73.1 & 74.0 & 50.0 & 71.0 & 62.7 \\
    \midrule
    GPT-4.1-2025-04-14 & 66.6 & 42.0 & 85.4 & 79.5 & 84.0 & 82.9 & 88.4 & 75.0 & 81.1 & 69.8 \\
    GPT-4o-2024-08-06 & 63.9 & 34.9 & 78.7 & 68.2 & 79.8 & 78.3 & 83.2 & 54.9 & 76.4 & 59.1 \\
    DeepSeek-V3-0324 & 69.4 & 54.7 & 81.5 & 76.6 & 80.6 & 81.2 & 84.7 & 68.5 & 79.1 & 70.3 \\
    \midrule
    \rowcolor[gray]{0.88}
    \Block{1-11}{\textit{Verifier Models}} & & & & & & & & & & \\
    \midrule
    xVerify-0.5B-I & 61.7 & 42.6 & 84.0 & 78.5 & 87.1 & 86.2 & 86.3 & 72.6 & 79.8 & 70.0 \\
    xVerify-8B-I & 64.3 & 42.6 & 84.3 & 78.9 & 86.1 & 85.1 & 88.7 & 74.9 & 80.8 & 70.4 \\
    xVerify-9B-C & 64.3 & 48.0 & 82.8 & 77.0 & 82.7 & 81.7 & 86.3 & 69.8 & 79.0 & 69.1 \\
    Tencent-Qwen2.5-7B-RLVR & 71.2 & 55.3 & 80.9 & 73.8 & 78.0 & 76.8 & 84.0 & 62.6 & 78.5 & 67.1 \\
    \midrule
    \rowcolor{blue!10}
    \Block{1-11}{\textit{CompassVerifiers}} & & & & & & & & & & \\
    \midrule 
    \rowcolor{blue!10}CompassVerifier-3B & 76.3 & 71.0 & 88.9 & 85.9 & 87.9 & 87.7 & 86.8 & 77.1 & 85.0 & 80.4 \\
    \rowcolor{blue!10}CompassVerifier-7B & 79.4 & 74.8 & 89.9 & 87.7 & 92.8 & 92.6 & 87.9 & 78.5 & 87.5 & 83.4 \\
    \rowcolor{blue!10}CompassVerifier-32B & \textbf{84.1} & \textbf{80.8} & \textbf{92.1} & \textbf{90.3} & \textbf{95.1} & \textbf{94.8} & \textbf{91.8} & \textbf{84.7} & \textbf{90.8} & \textbf{87.7} \\
    \bottomrule
    \end{NiceTabular}
    }
    \vspace{-1.0em}
\end{table*}
\begin{figure}[t]
    \centering
    \includegraphics[scale=0.25]{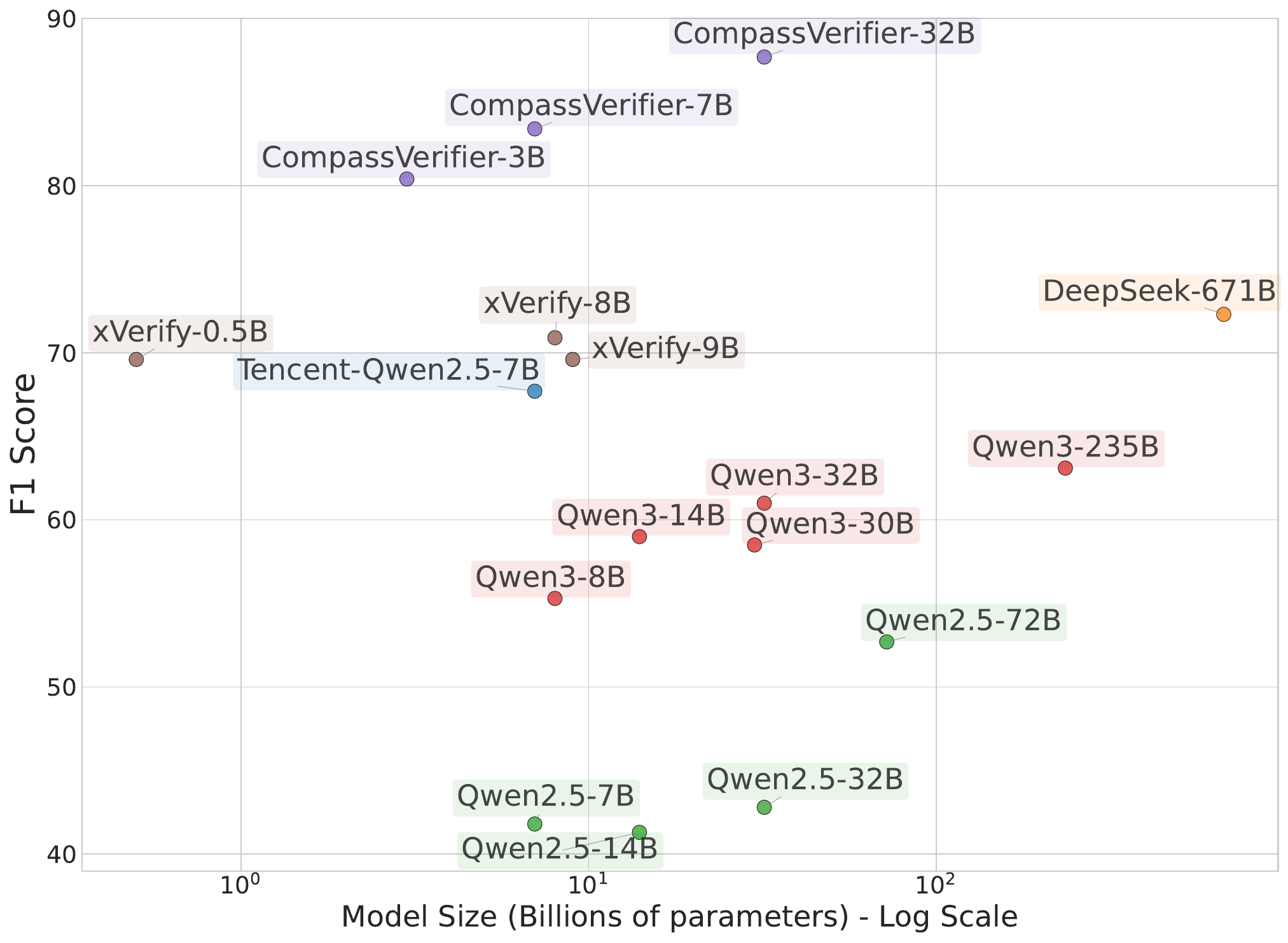}
    \caption{Model performances with size on VerifierBench. We show the F1 score in main results.}
    \label{fig_f1_vs_model_size_results}
\end{figure}
\paragraph{From the Perspective of the Domain.}
We show the main results of VerifierBench in \Cref{tab:main-results}.
Our CompassVerifier establishes new state-of-the-art performance across all VerifierBench categories, achieving 84.1--95.1\% accuracy and 80.8--94.8\% $F_1$-score in the 32B configuration. Three findings emerge: 1) As shown in \Cref{fig_f1_vs_model_size_results}, verification capability exhibits progressive improvement with increasing scale, demonstrating accuracy gains from \textbf{85.0\%} to \textbf{90.8\%} and $F_1$-score improvements from \textbf{80.4\%} to \textbf{87.7\%} as parameters scale from 3B to 32B. 2) Verification-specific architectures yield substantial gains: CompassVerifier-7B surpasses the similarly-sized original Qwen2.5-7B-Instruct by an absolute $F_1$-score improvement of 41.3\%. 
3) Despite progress, mathematical verification remains challenging (80.8\% best $F_1$-score vs. 94.8\% for knowledge), highlighting persistent gaps in stepwise logical validation. Our smallest 3B variant outperforms GPT-4.1 by an absolute $F_1$-score improvement of 10.6\%, demonstrating parameter efficiency. Consistent performance across domains further underscores the model's robustness. For instance, our CompassVerifier-32B model achieves high $F_1$-scores across all evaluated categories. Such consistency indicates a well-generalized verification capability, effectively handling diverse types of information and reasoning processes.

\begin{figure*}[t]
    \centering
    \begin{minipage}[b]{0.59\textwidth}
        \centering
        \begin{subfigure}[b]{0.495\textwidth}
            \centering
            \includegraphics[width=\linewidth]{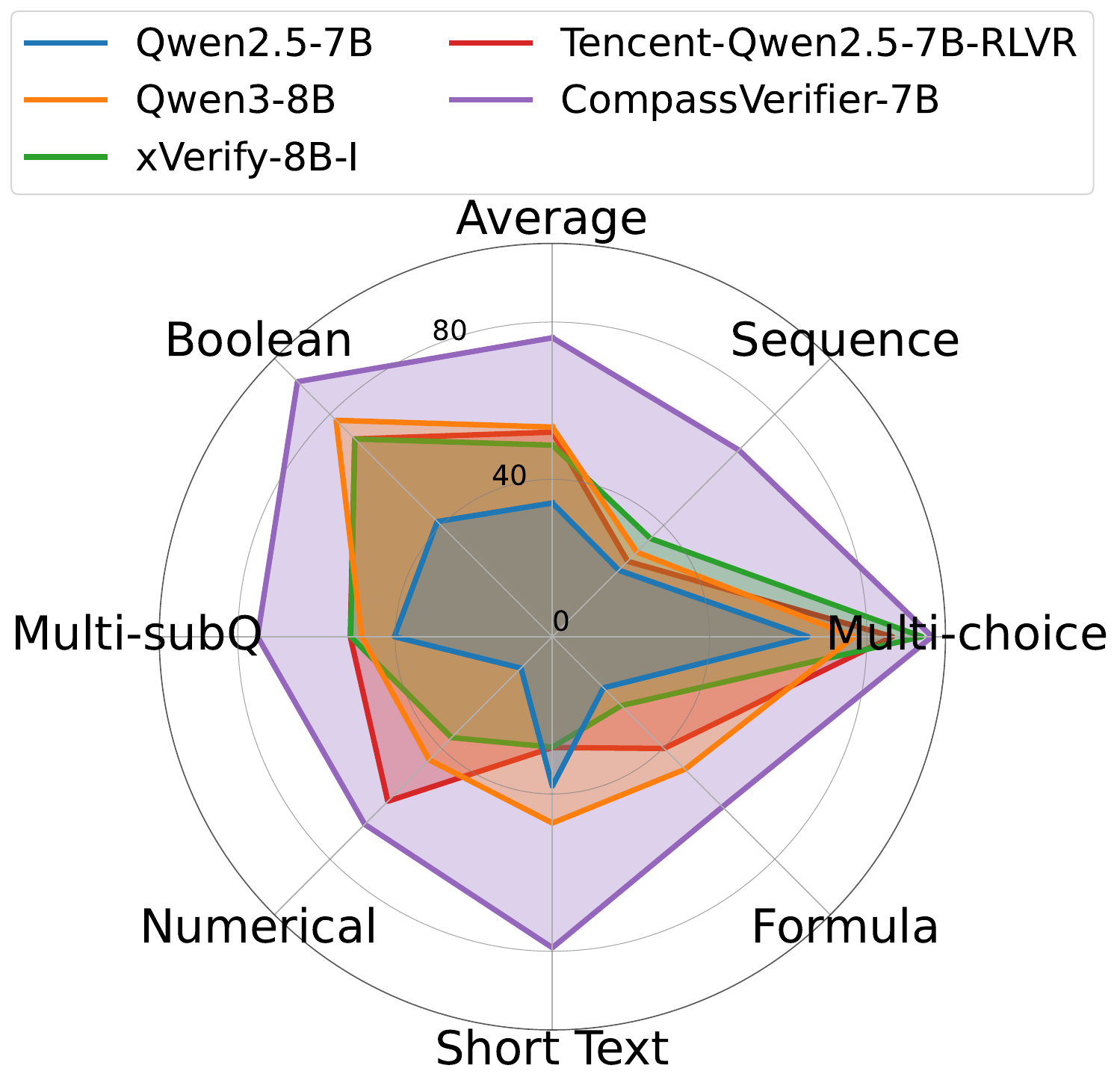}
            \caption{Seven answer types.}
            \label{fig_type_8b}
        \end{subfigure}
        \hfill
        \begin{subfigure}[b]{0.485\textwidth}
            \centering
            \includegraphics[width=\linewidth]{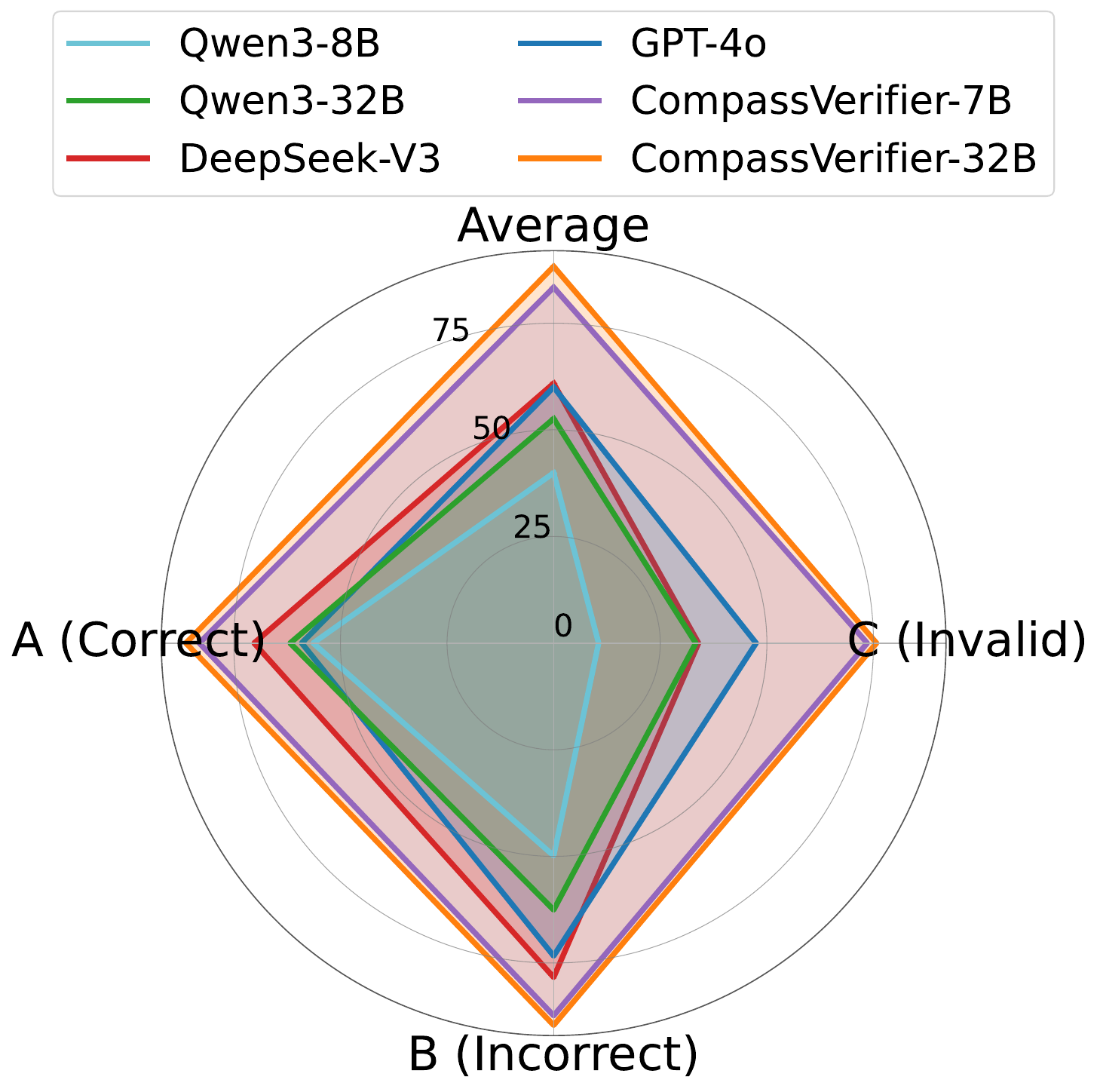}
            \caption{Ternary class labels.}
            \label{fig_3label}
        \end{subfigure}
        \vspace{-0.2cm}
        \caption{Results (F1) on VerifierBench across 7 answer types and 3 correctness labels.}
    \end{minipage}
    \hfill
    \begin{minipage}[b]{0.4\textwidth}
        \centering
        \includegraphics[width=0.8\linewidth]{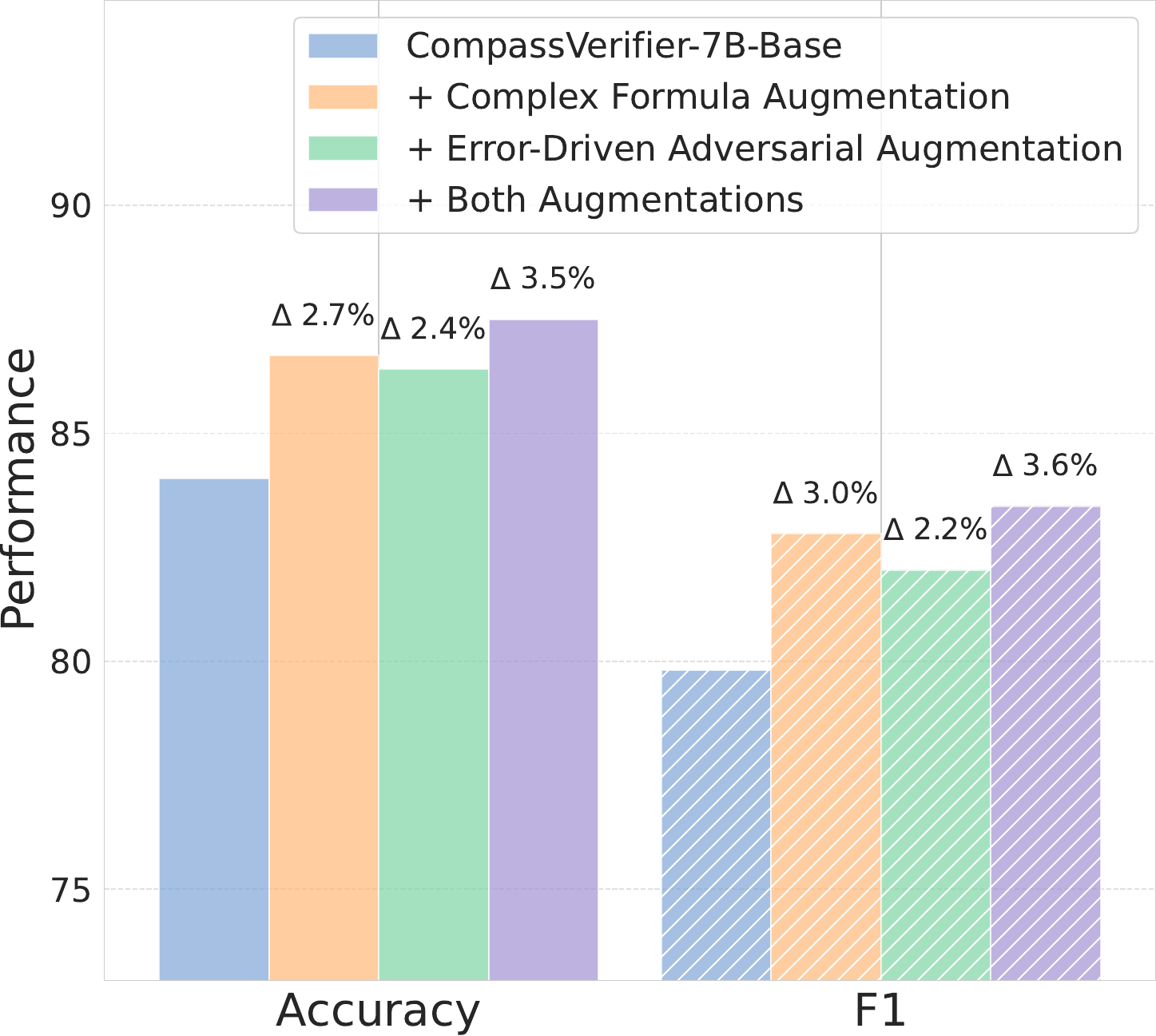}
        \vspace{-0.1cm}
        \caption{Ablation study on CompassVerifier-7B with different training technologies.}
        \label{fig_ablation_results}
    \end{minipage}
\end{figure*}

\vspace{-0.5em}
\paragraph{From the Perspective of the Answer Type.}
\Cref{fig_type_8b} demonstrates the performance comparison of similarly-sized models across different answer/question types. Notably, CompassVerifier-7B achieves consistent improvements across all categories. As evident from the results, multiple-choice questions emerge as the easiest category, with most models attaining strong performance, a finding attributable to their prevalence in evaluation benchmarks. However, baseline models show marked deficiencies in handling formula-based answers, multi-subquestions, and sequential answers, particularly struggling with sequential answers where none exceed 40 $F_1$-score. This likely stems from the inherent complexity of sequential answers, which often require element-by-element matching of multiple components, significantly increasing verification difficulty. These challenging cases represent precisely the focus of CompassVerifier and constitute critical directions for future research. The complete results are presented in \Cref{tab:qtype-results}.

\subsection{Analysis of CompassVerifier}

\paragraph{Beyond Binary Verification: Identifying Invalid Responses.}
\Cref{fig_3label} presents the three-class classification performance of six top-performing models. Notably, even advanced general LLMs like GPT-4o and DeepSeek-V3 without task-specific training exhibit significant performance bias, demonstrating substantially better results on categories A and B compared to C. Our manual analysis reveals that general models show particular insensitivity to duplicated patterns or truncated responses. To address this, we implemented a duplicate string detection script during data filtering (\Cref{{Data filtering pipeline}}). Crucially, we argue that Category C requires distinct treatment as they are particularly susceptible to reward hacking in RL training scenarios. Full results of the ternary classification performance are shown in \Cref{tab:3label-results}.

\paragraph{Impact of Data Augmentation Components.}
\Cref{fig_ablation_results} details the impact of our data augmentation strategies on CompassVerifier-7B. The baseline model (CompassVerifier-7B-Base) achieves 84.0\% accuracy and 79.8\% F1. 
Introducing \textit{Complex Formula Augmentation} alone improves accuracy to 86.7\% (+2.7) and F1 to 82.8\% (+3.0). This demonstrates the strategy's effectiveness in enhancing the model's capability to handle diverse formulaic expressions.
Similarly, \textit{Error-Driven Adversarial Augmentation} alone boosts accuracy to 86.4\% (+2.4) and F1 to 82.0\% (+2.2), underscoring its utility in fortifying the model against previously identified failure modes.
Combining both strategies yields the best performance, with accuracy reaching 87.5\% (+3.5) and F1 at 83.4\% (+3.6), demonstrating their complementary and synergistic contributions to overall verification capabilities. Details are shown in \Cref{tab:ablation}.

\paragraph{Generalization of CompassVerifier.}

To evaluate the generalization capability of CompassVerifier, we also conduct tests on the hard subset of VerifyBench~\citep{yan2025verifybench}, a recent concurrent work for benchmarking verification abilities. This subset primarily contains standard answers that involve long reasoning COT, making it particularly challenging to verify. Table \ref{tab:verify-bench} presents the performance comparison across different models. Here, ``Model-specific Prompt” indicates that xVerify/Tencent-RLVR employs their respective training prompts while other models use ours, whereas ``VerifyBench Prompt” denotes that all models utilize the same prompt provided with the VerifyBench dataset. Our analysis leads to the following findings: 1) CompassVerifier still outperforms both general LLMs of similar size, specialized verifier models, and even DeepSeek-V3; 2) Due to our \textbf{Generalizability Augmentation}, even under VerifyBench's prompt (deeper out-of-distribution setting), CompassVerifier maintains robust performance (score >86), while xVerify and Tencent-Qwen2.5-7B-RLVR completely fail to follow instructions.

\begin{table*}[t]
    \centering
    \caption{Performance on VerifyBench using different prompt strategies. We report Accuracy and F1 scores (\%) for both model-specific prompts and the standard VerifyBench prompts.}
    \label{tab:verify-bench}
    \resizebox{0.65\textwidth}{!}{
    \begin{NiceTabular}{lcccc}
    \toprule
    \multirow{2}{*}{Model} & 
    \multicolumn{2}{c}{Model-specific Prompt} & 
    \multicolumn{2}{c}{VerifyBench Prompt} \\
    \cmidrule(r){2-3} \cmidrule(l){4-5}
    & Acc & F1 & Acc & F1 \\
    \midrule
    \rowcolor[gray]{0.88}
    \Block{1-5}{\textit{General LLMs}} & & & & \\
    \midrule
    Qwen2.5-7B-Instruct & 65.4 & 39.8 & 60.9 & 45.0 \\
    Qwen2.5-32B-Instruct & 78.8 & 58.9 & 72.0 & 55.8 \\
    Qwen2.5-72B-Instruct & 78.5 & 61.7 & 63.0 & 50.0 \\
    DeepSeek-V3 & 81.8 & 62.2 & 78.6 & 60.9 \\
    \midrule
    \rowcolor[gray]{0.88}
    \Block{1-5}{\textit{Verifier Models}} & & & & \\
    \midrule
    xVerify-0.5B-I & 77.9 & 66.2 & - & - \\
    xVerify-8B-I & 83.2 & 70.7 & - & - \\
    xVerify-9B-C & 83.2 & 71.0 & - & - \\
    Tencent-Qwen2.5-7B-RLVR & 82.4 & 68.9 & - & - \\
    \midrule
    \rowcolor{blue!10}
    \Block{1-5}{\textit{CompassVerifiers}} & & & & \\
    \midrule
    \rowcolor{blue!10}CompassVerifier-3B & 87.4 & 77.4 & 86.2 & \textbf{75.0} \\
    \rowcolor{blue!10}CompassVerifier-7B & 88.1 & 79.0 & 86.0 & 73.3 \\
    \rowcolor{blue!10}CompassVerifier-32B & \textbf{89.7} & \textbf{81.1} & \textbf{86.8} & 74.3 \\
    \bottomrule
    \end{NiceTabular}
    }
    \vspace{-1.0em}
\end{table*}
\begin{figure}[t]
    \centering
    \includegraphics[width=.96\textwidth]{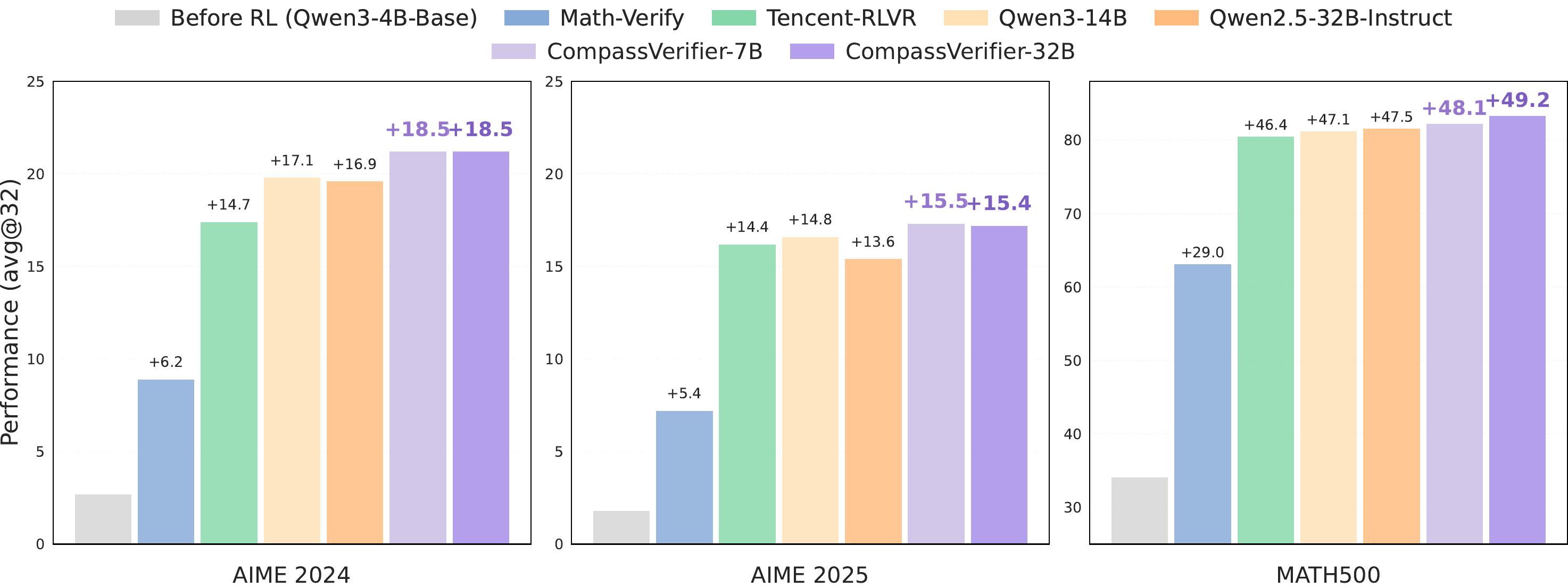}
    \caption{Experimental results of CompassVerifier as a reward model. We employ Math-Verify along with various general LLMs and verifier models as reward models for RL training, reporting the avg@32 performance on AIME24, AIME25, and MATH500.} \label{fig_rl_results}
\end{figure}

\subsection{CompassVerifier as Reward Model}

To validate the efficacy of CompassVerifier as a reward model in RL training, we examine its influence on enhancing the reasoning performance of models trained using RL. 
Specifically, we utilize GRPO \citep{shao2024deepseekmathpushinglimitsmathematical} to train base LLMs with rule-based verifier Math-Verify \citep{math-verify} and CompassVerifier and rigorously evaluate the reasoning capabilities of the trained models.
We use the challenging Open-S1~\citep{dang2025reinforcementlearningreasoningsmall} as the RL training corpus, which can also be considered an out-of-distribution dataset for CompassVerifiers. More experimental settings are provided in \Cref{app:details-verifier-as-reward}.


Comparative results are shown in \Cref{fig_rl_results} (Details in \Cref{tab_rl_results}). 
Experimental results demonstrate that models trained with CompassVerifier outperform the base model, surpass those trained with the rule-based verifier (Math-Verify), and exceed models using general LLMs or alternative verifiers as reward models. This highlights CompassVerifier's superior potential as a reward model, providing more precise evaluation for rollout trajectories generated in RL training.
Additionally, CompassVerifier's enhanced capacity to provide more effective signals (i.e., rewards) during training substantially improves the convergence efficiency of RL training.
The results also reveal a noticeable performance gap between rule-based and model-based verifiers. As the data types and disciplines covered by Reinforcement Learning from Verifiable Rewards (RLVR)~\citep{wang2025reinforcement} training continue to expand, rule-matching tools have become increasingly inadequate, which precisely motivated the development of CompassVerifier.
\section{Conclusion}
To address the critical gap in large-scale answer verification evaluation, we present \textbf{VerifierBench}, featuring a meticulously designed pipeline for large-scale data collection, filtering, and annotation. We also introduce \textbf{CompassVerifier}, a novel verification model specifically engineered to handle multi-domain scenarios, diverse answer types, varied prompt formats, and irregular responses. CompassVerifier achieves superior accuracy and robustness compared to larger general LLMs and baseline verifier models. We anticipate that VerifierBench and CompassVerifier would significantly advance research in answer verification for evaluation frameworks and reward modeling for RLVR.

\bibliographystyle{colm2025_conference}

\bibliography{colm/ref}


\clearpage
\onecolumn
\section{Appendix}
\subsection{Details of VerifierBench Statistics}
\label{Details of VerifierBench}
\begin{table}[h]
\centering
\caption{Dataset source distribution.}
\label{tab:VB_dataset}
\begin{tabular}{lcc}
\toprule
Source & Count & Percentage (\%) \\
\midrule
BBH & 639 & 22.68 \\
GaokaoBench & 201 & 7.14 \\
Math & 182 & 6.46 \\
MMLU Pro & 172 & 6.11 \\
GPQA Diamond & 51 & 1.81 \\
GSM8K & 14 & 0.50 \\
AIME2024 & 3 & 0.11 \\
SimpleQA & 97 & 3.44 \\
Numina Train & 106 & 3.76 \\
HLE & 355 & 12.60 \\
KorBench & 395 & 14.02 \\
OlympiadBench & 345 & 12.25 \\
ARC Prize Public Evaluation & 175 & 6.21 \\
TheoremQA & 82 & 2.91 \\
\bottomrule
\end{tabular}
\end{table}
\begin{table}[h]
\centering
\caption{Category distribution.}
\label{tab:VB_class}
\begin{tabular}{lcc}
\toprule
Category & Count & Percentage (\%) \\
\midrule
A & 1092 & 38.76 \\
B & 1526 & 54.17 \\
C & 199 & 7.06 \\
\bottomrule
\end{tabular}
\end{table}
\begin{table}[h]
\centering
\caption{Domain distribution.}
\label{tab:VB_domain}
\begin{tabular}{lcc}
\toprule
Domain & Count & Percentage (\%) \\
\midrule
General Reasoning & 1151 & 40.86 \\
Mathematical Reasoning & 900 & 31.95 \\
Knowledge & 387 & 13.74 \\
Scientific Reasoning & 379 & 13.45 \\
\bottomrule
\end{tabular}
\end{table}
\begin{table}[ht!]
\centering
\caption{Answer type distribution.}
\label{tab:VB_answer_type}
\begin{tabular}{lcc}
\toprule
Answer Type & Count & Percentage (\%) \\
\midrule
Multiple Choice & 891 & 31.63 \\
Short Text & 354 & 12.57 \\
Numerical & 434 & 15.41 \\
Formula & 343 & 12.18 \\
Multi-subproblem & 281 & 9.98 \\
Sequence & 468 & 16.61 \\
Boolean Answer & 46 & 1.63 \\
\bottomrule
\end{tabular}
\end{table}

\subsection{Details of VerifierBench Construction}
\textbf{Data Collection.} Our experimental evaluation encompasses a comprehensive collection of 53 LLMs, including representative examples such as  Qwen-2.5~\citep{yang2024qwen2}, LLaMA3~\citep{grattafiori2024llama}, DeepSeek-V3~\citep{liu2024deepseek}, DeepSeek-R1~\citep{guo2025deepseek}, GPT-4o~\citep{GPT-4o}, GPT-4o-mini~\citep{GPT-4o-mini}, Gemini~\citep{team2023gemini}, claude3-5~\citep{claude3-5}, Doubao-1.5-Pro~\citep{Doubao1-5-Pro}, InternLM~\citep{cai2024internlm2} and Mixtral~\citep{jiang2024mixtral}. All specific models are listed in Table \ref{tab_list_of_models}. These models are evaluated across sixteen diverse benchmarks: GSM8K~\citep{hosseini2024v}, Math~\citep{hendrycks2021measuring}, AIME2024~\citep{aime2024}, BBH~\citep{suzgun2022challenging}, GaokaoBench~\citep{zhang2023evaluating}, HLE~\citep{phan2025humanity}, KorBench~\citep{ma2024kor}, GPQA~\citep{rein2024gpqa}, SimpleQA~\citep{wei2024measuring}, ChineseSimpleQA~\citep{he2024chinese}, MMLU-Pro~\citep{wang2024mmlu}, ARC~\citep{chollet2024arc}, OlympiadBench~\citep{he-etal-2024-olympiadbench}, TheoremQA~\citep{chen-etal-2023-theoremqa}, NuminaMath~\citep{numina_math_datasets}, and Drop~\citep{Dua2019DROP}. Through the OpenCompass~\citep{2023opencompass} framework, we collected more than 1.32 million response models, creating the most comprehensive response datasets to date.

\noindent \textbf{VerifierBench Construction Details.}  
For samples with inconsistent verification results across multiple models and prompts, we identified numerous cases that were either redundant or unworthy of human annotation. We employed a string-matching script to detect and remove duplicate responses, which predominantly belonged to category C (invalid responses). Additionally, we utilized DeepSeek-V3 to identify problematic cases, including: (1) questions with obvious open-ended nature, (2) incomplete reference answers, and (3) proof-based problems - all of which cannot be objectively evaluated solely based on reference answers and may introduce ambiguity in test set evaluation. After deduplication, approximately 5,000 samples underwent human annotation, where annotators further flagged the aforementioned problematic types. Annotation results revealed that most of the inconsistent samples were ultimately labeled as category B (incorrect responses), suggesting a potential tendency of LLM judges toward false positives. To maintain better label balance, we further applied similarity-based filtering to remove redundant samples within the category B subset. This rigorous filtering process yielded a final high-quality dataset of 2,817 samples.

\subsection{Details of CompassVerifier Experiments}
\label{Details of CompassVerifier Experiments}
\noindent \textbf{Evaluation Setup.} We use OpenCompass~\citep{2023opencompass} and employ both F1 score and Accuracy as evaluation metrics, with particular emphasis on the F1 score, as it provides a more comprehensive assessment considering the precision, recall, and balance of the class distribution simultaneously. For all open-source models, we use vllm~\cite{kwon2023efficient} for the acceleration of inference. For all models, we employ temperature=1.0 for data synthesis and temperature=0.0 for evaluation/verification, with both $max\_gen\_len$ and $max\_model\_len$ set to their maximum values. We use the official prompt for Xverify and Tencent-Qwen2.5-7B-Instruct-RLVR, and a general non-cot prompt for CompassVerifier and general LLMs, which can be found in the first prompt in \Cref{Prompt_List}.

\noindent \textbf{Training Setup.}
We use XTuner~\citep{2023xtuner} for training our CompassVerifier model on Qwen2.5~\citep{yang2024qwen2} series models, largely adhering to the original hyperparameters. Fine-tuning is conducted using a learning rate of \(2 \times 10^{-5}\) with a max sequence length 32768. A multiplicative learning rate decay is applied after each epoch, with a gamma value of \(0.85\). The batch sizes are set to 32. All models are trained for one epoch on the training set and fully fine-tuned on 8×A100 80GB GPUs.

\begin{table*}[t]
    \centering
    \caption{Detailed results on VerifierBench across different question types. We report Accuracy (Acc.) and F1 scores (\%) for various problem categories and their average. Bold numbers indicate the best performance in each column.}
    \label{tab:qtype-results}
    \resizebox{0.99\textwidth}{!}{
    \begin{NiceTabular}{lcccccccccccccc|cc}
    \toprule
    \multirow{2}{*}{Model} &
    \multicolumn{2}{c}{Boolean} &
    \multicolumn{2}{c}{Multi-sub} &
    \multicolumn{2}{c}{Numerical} &
    \multicolumn{2}{c}{Short Text} &
    \multicolumn{2}{c}{Formula} &
    \multicolumn{2}{c}{Multi-choice} &
    \multicolumn{2}{c}{Sequence} &
    \multicolumn{2}{|c}{Average} \\
    \cmidrule(r){2-3} \cmidrule(r){4-5} \cmidrule(r){6-7} \cmidrule(r){8-9} \cmidrule(r){10-11} \cmidrule(r){12-13} \cmidrule(r){14-15} \cmidrule(l){16-17}
    & Acc. & F1 & Acc. & F1 & Acc. & F1 & Acc. & F1 & Acc. & F1 & Acc. & F1 & Acc. & F1 & Acc. & F1 \\
    \midrule
    \rowcolor[gray]{0.88}
    \Block{1-17}{\textit{General LLMs}} & & & & & & & & & & & & & & & & \\
    \midrule
    Qwen2.5-7B-Instruct & 63.0 & 41.4 & 45.9 & 40.2 & 49.5 & 11.3 & 65.0 & 38.0 & 53.5 & 18.4 & 62.0 & 65.0 & 59.2 & 23.9 & 56.9 & 34.0 \\
    Qwen2.5-14B-Instruct & 63.0 & 66.7 & 54.5 & 45.0 & 57.4 & 39.3 & 59.9 & 42.3 & 53.8 & 26.9 & 49.0 & 45.9 & 68.8 & 34.8 & 58.0 & 43.0 \\
    Qwen2.5-32B-Instruct & 58.7 & 53.7 & 65.8 & 37.7 & 56.7 & 33.9 & 61.3 & 27.7 & 59.3 & 19.5 & 55.8 & 52.5 & 80.6 & 19.5 & 62.6 & 34.9 \\
    Qwen2.5-72B-Instruct & 73.9 & 71.4 & 65.8 & 46.7 & 62.0 & 36.8 & 57.9 & 47.7 & 57.0 & 27.5 & 61.9 & 62.4 & 74.8 & 40.4 & 64.8 & 47.6 \\
    \midrule
    Qwen3-8B & 73.9 & 77.8 & 50.2 & 48.5 & 52.5 & 44.3 & 52.3 & 47.4 & 54.7 & 47.7 & 70.4 & 76.8 & 53.0 & 30.4 & 58.1 & 53.3 \\
    Qwen3-14B & 69.6 & 66.7 & 69.8 & 52.0 & 64.8 & 39.0 & 76.6 & 56.1 & 66.6 & 27.7 & 72.4 & 73.8 & 84.6 & 39.0 & 72.0 & 50.6 \\
    Qwen3-30B-A3B & 71.7 & 69.8 & 45.9 & 44.9 & 66.1 & 66.4 & 53.7 & 47.4 & 48.8 & 51.4 & 74.9 & 79.8 & 55.1 & 28.1 & 59.5 & 55.4 \\
    Qwen3-32B & 80.4 & 80.9 & 63.4 & 55.9 & 64.8 & 51.4 & 68.6 & 57.1 & 64.2 & 44.3 & 74.3 & 77.8 & 78.4 & 46.0 & 70.6 & 59.1 \\
    Qwen3-235B-A22B & 67.4 & 57.1 & 60.9 & 52.6 & 63.8 & 48.9 & 67.8 & 56.1 & 62.5 & 43.5 & 79.0 & 82.6 & 83.3 & 50.4 & 69.2 & 55.9 \\
    \midrule
    GPT-4.1-2025-04-14 & 80.4 & 80.0 & 68.3 & 44.7 & 64.1 & 31.6 & 83.1 & 64.7 & 68.6 & 22.9 & 89.4 & 91.0 & 88.3 & 43.3 & 77.4 & 54.0 \\
    GPT-4o-2024-08-06 & 65.2 & 63.6 & 63.7 & 37.0 & 63.6 & 29.5 & 79.7 & 54.4 & 67.2 & 11.0 & 80.0 & 81.9 & 86.8 & 35.4 & 72.3 & 44.7 \\
    DeepSeek-V3-0324 & 63.0 & 56.4 & 61.2 & 52.0 & 68.2 & 48.9 & 81.6 & 66.3 & 69.5 & 39.3 & 85.4 & 87.6 & 85.5 & 54.1 & 73.5 & 57.8 \\
    \midrule
    \rowcolor[gray]{0.88}
    \Block{1-17}{\textit{Verifier Models}} & & & & & & & & & & & & & & & & \\
    \midrule
    xVerify-0.5B-I & 67.4 & 59.5 & 66.9 & 25.6 & 63.6 & 37.8 & 64.7 & 36.6 & 60.8 & 22.0 & 95.7 & 96.6 & 85.5 & 35.0 & 72.1 & 44.7 \\
    xVerify-8B-I & 71.7 & 71.1 & 73.0 & 51.3 & 65.2 & 36.3 & 65.3 & 28.1 & 66.6 & 24.8 & 92.6 & 94.0 & 88.3 & 35.3 & 74.7 & 48.7 \\
    xVerify-9B-C & 67.4 & 70.6 & 76.9 & 50.4 & 65.2 & 40.8 & 58.8 & 34.8 & 63.4 & 30.0 & 92.3 & 93.6 & 85.9 & 29.8 & 72.8 & 50.0 \\
    Tencent-Qwen2.5-7B-Instruct-RLVR & 71.7 & 71.1 & 69.0 & 51.4 & 74.9 & 59.2 & 71.2 & 28.2 & 69.8 & 40.2 & 84.2 & 86.5 & 85.0 & 27.1 & 75.1 & 52.0 \\
    \midrule
    \rowcolor{blue!10}
    \Block{1-17}{\textit{CompassVerifiers}} & & & & & & & & & & & & & & & & \\
    \midrule
    \rowcolor{blue!10} CompassVerifier-3B & 87.0 & 86.4 & 80.8 & 69.3 & 75.8 & 65.1 & 78.8 & 59.9 & 68.8 & 57.4 & 95.7 & 96.6 & 87.6 & 52.5 & 82.1 & 69.3 \\
    \rowcolor{blue!10} CompassVerifier-7B & 91.3 & 91.7 & 85.1 & 75.0 & 77.0 & 67.5 & 87.6 & 79.1 & 71.1 & 61.2 & 95.6 & 96.6 & 90.2 & 67.1 & 85.4 & 76.0\\
    \rowcolor{blue!10} CompassVerifier-32B & \textbf{95.7} & \textbf{95.8} & \textbf{93.6} & \textbf{89.2} & 80.9 & 74.7 & \textbf{88.4} & \textbf{79.8} & 79.9 & \textbf{71.4} & 96.2 & 97.0 & \textbf{93.2} & \textbf{74.6} & \textbf{89.2} & \textbf{83.0} \\
    \bottomrule
    \end{NiceTabular}
    }
    \vspace{-1.0em}
\end{table*}
\begin{table*}[t]
    \centering
    \caption{Three-label classification performance on VerifierBench. Beyond binary correctness (correct/incorrect), this evaluation requires models to identify invalid responses. We report Accuracy and macro-$F_1$ scores (in \%) across four distinct categories and their overall average.}
    \label{tab:3label-results}
    \resizebox{0.99\textwidth}{!}{
    \begin{NiceTabular}{lcccccccc|cc}
    \toprule
    \multirow{2}{*}{Model} &
    \multicolumn{2}{c}{Math} &
    \multicolumn{2}{c}{General Reasoning} &
    \multicolumn{2}{c}{Knowledge} &
    \multicolumn{2}{c}{Science} &
    \multicolumn{2}{c}{Average} \\
    \cmidrule(r){2-3} \cmidrule(r){4-5} \cmidrule(r){6-7} \cmidrule(r){8-9} \cmidrule(l){10-11}
     & Acc. & macro-F1 & Acc. & macro-F1 & Acc. & macro-F1 & Acc. & macro-F1 & Acc. & macro-F1 \\
    \midrule
    \rowcolor[gray]{0.88}
    \Block{1-11}{\textit{General LLMs}} & & & & & & & & & & \\
    \midrule
    Qwen2.5-7B-Instruct & 39.6 & 29.2 & 49.2 & 37.8 & 45.2 & 34.6 & 50.3 & 34.2 & 46.1 & 34.0 \\
    Qwen2.5-14B-Instruct & 44.2 & 37.7 & 50.9 & 40.1 & 42.9 & 37.6 & 57.1 & 44.1 & 48.8 & 39.9 \\
    Qwen2.5-32B-Instruct & 46.0 & 35.7 & 59.8 & 47.8 & 55.6 & 45.7 & 70.8 & 52.5 & 58.0 & 45.4 \\
    Qwen2.5-72B-Instruct & 51.1 & 43.0 & 57.3 & 48.6 & 67.4 & 52.2 & 72.9 & 58.8 & 62.2 & 50.7 \\
    \midrule
    Qwen3-8B & 48.2 & 35.8 & 54.0 & 42.3 & 56.1 & 41.1 & 47.9 & 36.5 & 51.5 & 38.9 \\
    Qwen3-14B & 61.3 & 57.3 & 72.3 & 63.5 & 65.4 & 54.7 & 74.7 & 61.9 & 68.4 & 59.4 \\
    Qwen3-30B & 53.3 & 45.6 & 49.6 & 42.1 & 54.8 & 50.2 & 45.0 & 39.0 & 50.7 & 44.2 \\
    Qwen3-32B & 57.2 & 54.2 & 61.6 & 54.4 & 60.2 & 51.7 & 58.7 & 50.0 & 59.4 & 52.6 \\
    Qwen3-235B-A22B & 58.8 & 42.8 & 73.8 & 55.0 & 65.4 & 48.6 & 67.6 & 52.4 & 66.4 & 49.7 \\
    \midrule
    GPT-4.1-2025-04-14 & 61.7 & 59.6 & 78.1 & 73.6 & 78.3 & 69.7 & 79.5 & 68.4 & 74.4 & 67.8 \\
    GPT-4o-2024-08-06 & 57.9 & 53.9 & 68.3 & 62.9 & 73.4 & 66.0 & 71.1 & 57.1 & 67.7 & 60.0 \\
    DeepSeek-V3-0324 & 63.2 & 49.1 & 77.4 & 66.2 & 76.5 & 60.3 & 80.5 & 67.8 & 74.4 & 60.9 \\
    \midrule
    \rowcolor{blue!10}
    \Block{1-11}{\textit{CompassVerifiers}} & & & & & & & & & & \\
    \midrule    
    \rowcolor{blue!10}CompassVerifier-3B & 73.4 & 68.8 & 87.4 & 85.6 & 86.3 & 87.1 & 87.6 & 80.8 & 83.7 & 80.6 \\
    \rowcolor{blue!10}CompassVerifier-7B & 77.7 & 74.3 & 88.1 & 87.6 & 91.5 & 92.6 & 86.0 & 79.1 & 85.8 & 83.4 \\
    \rowcolor{blue!10}CompassVerifier-32B & \textbf{82.0} & \textbf{79.6} & \textbf{90.0} & \textbf{90.7} & \textbf{94.3} & \textbf{95.9} & \textbf{91.3} & \textbf{86.8} & \textbf{89.4} & \textbf{88.3} \\
    \bottomrule
    \end{NiceTabular}
    }
\end{table*}
\begin{table*}[t]
  \centering
  \caption{Ablation study on CompassVerifier-7B with different augmentation strategies on VerifierBench main results. \textit{Complex Formula Augmentation} enhances formula variants verification, \textit{Error-Driven Adversarial Augmentation} fortifies against failure cases.}
  \label{tab:ablation}
  \resizebox{0.9 \textwidth}{!}{
  \begin{tabular}{lcccc}
    \toprule
    \textbf{Setting} & \textbf{Accuracy (\%)} & \textbf{$\Delta$ Acc (\%)} & \textbf{F1 (\%)} & \textbf{$\Delta$ F1 (\%)} \\
    \midrule
    CompassVerifier-7B-Base                         & 84.0  & -    & 79.8   & -   \\
    \quad + Complex Formula Augmentation            & 86.7  & +2.7 & 82.8  & +3.0 \\
    \quad + Error-Driven Adversarial Augmentation   & 86.4  & +2.4 & 82.0  & +2.2 \\
    \quad + Both Augmentations                      & \textbf{87.5}  & \textbf{+3.5} & \textbf{83.4}  & \textbf{+3.6} \\
    \bottomrule
  \end{tabular}
}
\end{table*}
\clearpage
\begin{table}[t]
    \centering
    \caption{Experimental results of CompassVerifier as a reward model. We report the avg@32 performance on AIME24, AIME25, and MATH500.} \label{tab_rl_results}
    \resizebox{.8\textwidth}{!}{
    \begin{tabular}{lccc}
        \toprule
        \textbf{Model} & \textbf{AIME24} & \textbf{AIME25} & \textbf{MATH500} \\
        \midrule
        \rowcolor[gray]{0.88}
        \multicolumn{4}{c}{\textit{Original Model Performance}} \\
        \midrule
        Qwen3-4B-Base & 2.7 & 1.8 & 34.1 \\
        \midrule
        \rowcolor[gray]{0.88}
        \multicolumn{4}{c}{\textit{RL with Rule-based Verifier}} \\
        \midrule
        Math-Verify & 8.9 & 7.2 & 63.1 \\
        \midrule
        \rowcolor[gray]{0.88}
        \multicolumn{4}{c}{\textit{RL with Model-based Verifier}} \\
        \midrule
        Tencent-RLVR & 17.4 & 16.2 & 80.5 \\
        Qwen3-14B & 19.8 & 16.6 & 81.2 \\
        Qwen2.5-32B & 19.6 & 15.4 & 81.6 \\
        \rowcolor{blue!10} CompassVerifier-7B & \textbf{21.2} & \textbf{17.3} & 82.2 \\
        \rowcolor{blue!10} CompassVerifier-32B & \textbf{21.2} & 17.2 & \textbf{83.3} \\
        \bottomrule
    \end{tabular}
    } 
    \vspace{-1.0em}
\end{table}
\subsection{Details of Meta Error Patterns}
\label{Details of Meta Error Patterns}

We display the meta error patterns in three categories: A (Correct), B (Incorrect), and C (Invalid) as shown in the following figures.


\begin{promptbox}{Meta Pattern: A (Correct)}
\begin{itemize}[noitemsep,leftmargin=*]
\item The units in the LLM Response differ from those in the final answer, resulting in different numerical expressions, but they are consistent upon conversion, should be judged as Correct.

\item The reference answer is an extremely complex formula, and the LLM Response appears very different in form but simplifies to an equivalent expression, with no explicit requirement for simplification in the question, should be judged as Correct.

\item The question requires calculating a numerical decrease, and the LLM Response has the opposite sign of the reference answer because either uses negative signs to represent decrease, but they are equivalent, should be judged as Correct.

\item The reference answer provides multiple candidate answers without requiring all possibilities. The LLM Response provides one of them, should be judged as Correct.

\item The question doesn't explicitly specify answer format (numerical or formula). The LLM Response and reference answer differ in form but are equivalent when calculated, should be judged as Correct.
\item The question requires specific formatting (order, capitalization, etc.). While the LLM Response appears different from the reference answer in formatting, upon inspection it fully complies, should be judged as Correct.
\item When calculating values with units, the reference answer and LLM Response may differ in unit representation or numerical values, but are equivalent after unit conversion, should be judged as Correct.
\item For multiple-choice or true/false questions, the LLM Response ultimately gives the correct answer despite showing significant uncertainty, should be judged as Correct.
\item The question requires expressions meeting simple conditions (sum, product, logical relations, etc.), and the reference answer may include multiple valid forms. The LLM Response differs in form but meets all requirements, should be judged as Correct.
\item The LLM initially provides an incorrect answer but corrects it after reflection, should be judged as Correct.
\item The reference answer consists of multiple sub-questions. The LLM answers all sub-questions correctly during reasoning, even if not presented together at the end, should be judged as Correct.
\end{itemize}
\end{promptbox}

\begin{promptbox}{Meta Pattern: B (Inorrect)}
\begin{itemize}[noitemsep,leftmargin=*]
\item For multiple-choice questions, the LLM Response selects the correct option but follows with unrelated option content, should be judged as Incorrect.
\item The question requires formula simplification. The LLM answer isn't fully simplified to minimal form, even if equivalent to the reference answer, should be judged as Incorrect.
\item The reference answer is a formula with specified output format. The LLM answer doesn't comply with this format, even if equivalent, should be judged as Incorrect.
\item The question requires an expression where the sum equals a certain value with each number used once. The LLM Response repeats numbers while satisfying the sum, should be judged as Incorrect.
\item The reference answer is an un-simplified logical formula after substitution. The LLM Response is incorrect due to simplification causing format errors, should be judged as Incorrect.
\item The LLM Response only provides solution code without final results, should be judged as Incorrect.
\item The LLM Response (formula/numerical) and reference answer aren't equivalent when calculated, should be judged as Incorrect.
\item When describing numerical intervals, the reference answer and LLM Response differ in endpoint inclusion (open/closed), should be judged as Incorrect.
\item For sequence decryption requiring exact matching, the LLM Response doesn't match the reference answer, should be judged as Incorrect.
\item The reference answer is a long sequence requiring exact correspondence. The LLM Response has minor differences with some errors, should be judged as Incorrect.
\item The question explicitly requires multiple candidate answers (provided in reference), but the LLM Response gives only one, should be judged as Incorrect.
\item The LLM initially provides a correct answer but changes to incorrect or "unanswerable" after reflection, should be judged as Incorrect.
\item For symbolic sequences, the LLM Response contains garbled characters, should be judged as Incorrect.
\item The reference answer is numerical, and the LLM Response provides more decimal places but rounds differently, should be judged as Incorrect.
\item The reference answer is an extremely large number, and the LLM Response provides a high-order power expression that doesn't match after calculation, should be judged as Incorrect.
\item After detailed reasoning, the LLM Response fails to provide a clear answer or states the question is unanswerable, should be judged as Incorrect.
\item For multi-part questions, the number of final answers in the LLM Response doesn't match the reference answer, should be judged as Incorrect.
\end{itemize}
\end{promptbox}

\begin{promptbox}{Meta Pattern: C (Invalid)}
\begin{itemize}[noitemsep,leftmargin=*]
\item The question contains multiple sub-questions, but the number of reference answers doesn't match, indicating quality issues, should be judged as Invalid.
\item The reference answer has serious omissions, truncation, or formatting issues, should be judged as Invalid.
\item The question itself has serious omissions, truncation, or formatting issues, should be judged as Invalid.
\item The LLM doesn't answer normally, stating it needs more information or internet access, should be judged as Invalid.
\item The LLM Response is clearly truncated and incomplete, should be judged as Invalid.
\item The LLM Response is mostly garbled text with no valuable information extractable, should be judged as Invalid.
\item The LLM Response contains extensive meaningless repetition, making correct answers unidentifiable, should be judged as Invalid.
\end{itemize}
\end{promptbox}

\subsection{Meta-Judge Template Generation Fields}
\begin{table}[ht]
\centering
\caption{Meta-Judge Template Generation Fields (Academic Disciplines and Subfields)}
\label{tab:template_fields}
\resizebox{\textwidth}{!}{
\scriptsize
\begin{tabular}{ll>{\raggedright\arraybackslash}p{8cm}}
\toprule
\textbf{Category} & \textbf{Discipline} & \textbf{Subfields} \\
\midrule
\multirow{6}{*}{Natural Sciences} 
& Mathematics & Differential calculus, Integral calculus, Probability statistics, Operations research, Mathematical logic, Financial mathematics, Topology, Algebraic geometry \\
& Physics & Theoretical physics, Quantum mechanics, Condensed matter physics, Astrophysics, Nuclear physics, Optics, Acoustics \\
& Chemistry & Analytical chemistry, Organic chemistry, Inorganic chemistry, Physical chemistry, Materials chemistry, Environmental chemistry, Chemical biology \\
& Biology & Molecular biology, Genetics, Ecology, Cell biology, Biochemistry, Microbiology \\
& Earth Sciences & Geology, Geophysics, Atmospheric sciences, Oceanography, Environmental science, Paleontology \\
& Statistics & Data science, Biostatistics, Economic statistics, Machine learning algorithms, Bayesian analysis \\

\midrule
\multirow{10}{*}{Engineering}
& Mechanical Engineering & Mechanical design \& manufacturing, Automatic control, Robotics, Vehicle engineering, Thermal \& power engineering, MEMS \\
& Computer Science \& Technology & Artificial intelligence, Computer networks, Software engineering, Computer vision, Cybersecurity, Big data analytics \\
& Electronic Information Engineering & Communication engineering, IC design, Optoelectronic technology, Wireless sensor networks, Smart grid \\
& Civil Engineering & Structural engineering, Bridge \& tunnel design, Geotechnical engineering, Hydraulic engineering, Urban planning \\
& Materials Science \& Engineering & Nanomaterials, Metallic materials, Polymer materials, Composite materials, Material processing \\
& Chemical Engineering & Chemical process design, Petroleum refining, Biochemical engineering, Catalytic reaction engineering, Separation technology \\
& Environmental Engineering & Pollution control technology, Environmental monitoring, Ecological restoration, Solid waste treatment, Clean energy development \\
& Aerospace Engineering & Aircraft design, Propulsion systems, Aerodynamics, Satellite navigation, Aerospace materials \\
& Biomedical Engineering & Medical imaging technology, Biomaterials, Artificial organs, Biosensors, Rehabilitation engineering \\
& Energy \& Power Engineering & Nuclear technology, Wind energy development, Solar energy utilization, Fuel cells, Thermal system optimization \\
\bottomrule
\end{tabular}}
\end{table}

\clearpage
\subsection{Cases in VerifierBench}
\label{Cases_VerifierBench}
\begin{promptbox}{Example I: Verification as Correct (Gold Judgment: A)}
\textcolor{blue}{\textbf{Question}}  
Let $f(x) = \sqrt{x}$ and $g(x) = x^2.$ Find $f(g(f(g(f(8))))).$

\textcolor{blue}{\textbf{Gold Answer}}  
$2\sqrt{2}$

\textcolor{blue}{\textbf{LLM Response}}  
To solve the problem $f(g(f(g(f(8)))))$, we need to evaluate the functions step by step, starting from the innermost function and working our way out.

1. Evaluate $f(8)$:
   \[ f(x) = \sqrt{x} \]
   \[ f(8) = \sqrt{8} \]

2. Evaluate $g(f(8))$:
   \[ g(x) = x^2 \]
   \[ g(\sqrt{8}) = (\sqrt{8})^2 = 8 \]

3. Evaluate $f(g(f(8)))$:
   \[ f(8) = \sqrt{8} \]
   \[ g(\sqrt{8}) = 8 \]
   \[ f(8) = \sqrt{8} \]

4. Evaluate $g(f(g(f(8))))$:
   \[ f(8) = \sqrt{8} \]
   \[ g(\sqrt{8}) = 8 \]
   \[ f(8) = \sqrt{8} \]
   \[ g(\sqrt{8}) = 8 \]

5. Evaluate $f(g(f(g(f(8)))))$:
   \[ f(8) = \sqrt{8} \]
   \[ g(\sqrt{8}) = 8 \]
   \[ f(8) = \sqrt{8} \]
   \[ g(\sqrt{8}) = 8 \]
   \[ f(8) = \sqrt{8} \]

After evaluating the functions step by step, we find that:
\[ f(g(f(g(f(8))))) = \sqrt{8} \]

Thus, the final answer is:
\[ \boxed{\sqrt{8}} \]

\textcolor{blue}{\textbf{Gold Judgment}}  
A (Correct)

\textcolor{blue}{\textbf{Meta Data}}  
\begin{itemize}
  \item Data Source: math
  \item Domain: Math
  \item Answer Type: Numerical
\end{itemize}
\end{promptbox}

\begin{promptbox}{Example II: Verification as Incorrect (Gold Judgment: B)}
\textcolor{blue}{\textbf{Question}}
Sort the following words alphabetically:

List: marshmallow doge alpenstock ocean accept angus drool jutish resistive chromium vociferous castigate prize octennial stonewort concision elizabethan

\textcolor{blue}{\textbf{Gold Answer}}
accept alpenstock angus castigate chromium concision doge drool elizabethan jutish marshmallow ocean octennial prize resistive stonewort vociferous

\textcolor{blue}{\textbf{LLM Response}}
So the answer is accept, alpenstock, angus, castigate, chromium, concision, doge, drool, elizabethan, jutish, marshmallow, octennial, ocean, prize, resistive, stonewort, vociferous.

\textcolor{blue}{\textbf{Gold Judgment}}
B (Incorrect)

\textcolor{blue}{\textbf{Meta Data}}
\begin{itemize}
    \item Data Source: BBH/bbh-word\_sorting
    \item Domain: General Reasoning
    \item Answer Type: Sequence
\end{itemize}
\end{promptbox}

\begin{promptbox}{Example III: Verification as Invalid (Gold Judgment: C)}
\textcolor{blue}{\textbf{Question}}
A droplet of molten lead of average 1/8 inch diameter, at 750\textdegree F, falls from a height of 60 ft, and solidifies by the cooling effect of the surrounding air. The surrounding temperature is 70\textdegree F. If the lead solidifies at 621\textdegree F, calculate the coefficient of heat transfer. 

Given properties for lead:
\begin{itemize}
    \item $C_p$ = 0.031 Btu/lbm-\textdegree F
    \item $\rho$ = 710 lbm/ft$^3$
    \item $h_{fusion}$ = 10.6 Btu/lbm
\end{itemize}

Options:
\begin{itemize}
    \item A. 59.2 Btu/hr-ft$^2$-\textdegree F
    \item B. 45.7 Btu/hr-ft$^2$-\textdegree F
    \item C. 52.8 Btu/hr-ft$^2$-\textdegree F
    \item D. 63.4 Btu/hr-ft$^2$-\textdegree F
\end{itemize}

\textcolor{blue}{\textbf{Gold Answer}}
A

\textcolor{blue}{\textbf{LLM Response}}
To solve this problem, we need to calculate the coefficient of heat transfer (h) using the formula:

Q = h * A * $\Delta$T

where:
- Q is the heat transferred (in Btu)
- A is the surface area of the droplet (in ft$^2$)
- $\Delta$T is the temperature difference (in \textdegree F)

First, let's find the volume (V) of the droplet using the given average diameter (1/8 inch):

V = $\pi$ * (d/2)$^3$
V = $\pi$ * (1/16)$^3$
V = $\pi$ * (1/4096) ft$^3$

Now, let's find th

\textcolor{blue}{\textbf{Gold Judgment}}
C (Invalid)

\textcolor{blue}{\textbf{Meta Data}}
\begin{itemize}
    \item Data Source: mmlu\_pro/mmlu\_pro\_engineering
    \item Domain: Science
    \item Answer Type: Multiple Choice
\end{itemize}
\end{promptbox}

\clearpage
\subsection{Prompt List}
\label{Prompt_List}
\begin{lstlisting}[caption={Prompt for general LLM evaluation }]
Please as a grading expert, judge whether the final answers given by the candidates below are consistent with the standard answers, that is, whether the candidates answered correctly. 
Here are some evaluation criteria:
1. Please refer to the given standard answer. You don't need to re-generate the answer to the question because the standard answer has been given. You only need to judge whether the candidate's answer is consistent with the standard answer according to the form of the question. THE STANDARD ANSWER IS ALWAYS CORRECT AND THE QUESTION IS PERFECTLY VALID. NEVER QUESTION THEM.
2. ONLY compare the FINAL ANSWER - COMPLETELY IGNORE any potential errors in the REASONING PROCESSES.
3. Some answers may be expressed in different ways, such as some answers may be a mathematical expression, some answers may be a textual description, as long as the meaning expressed is the same. Before making a judgment, please understand the question and the standard answer first, and then judge whether the candidate's answer is correct.
4. Some answers may consist of multiple items, such as multiple-choice questions, multiple-select questions, fill-in-the-blank questions, etc. Regardless of the question type, the final answer will be considered correct as long as it matches the standard answer, regardless of whether the reasoning process is correct. For multiple-select questions and multi-blank fill-in-the-blank questions, all corresponding options or blanks must be answered correctly and match the standard answer exactly to be deemed correct.
5. If the prediction is given with \\boxed{{}}, please ignore the \\boxed{{}} and only judge whether the candidate's answer is consistent with the standard answer.
6. If the candidate's answer is invalid (e.g., incomplete (cut off mid-response), lots of unnormal repetitive content, or irrelevant to the question, saying it can't answer the question because some irresistible factors, like ethical issues, no enough information, etc.), select option C (INVALID).Please judge whether the following answers are consistent with the standard answer based on the above criteria. Grade the predicted answer of this new question as one of:
A: CORRECT 
B: INCORRECT
C: INVALID
Just return the letters "A", "B", or "C", with no text around it.
Here is your task. Simply reply with either CORRECT, INCORRECT, or INVALID. Don't apologize or correct yourself if there was a mistake; we are just trying to grade the answer.
<Original Question Begin>:
{question}
<Original Question End>
<Standard Answer Begin>:
{gold_answer}
<Standard Answer End>
<Candidate's Answer Begin>: 
{llm_response}
<Candidate's Answer End>
Judging the correctness of the candidate's answer:
\end{lstlisting}

\begin{lstlisting}[caption={Prompt A for CoT answer verification}]
As a grading expert, your task is to determine whether the candidate's final answer matches the provided standard answer. Follow these evaluation guidelines precisely:

Evaluation Protocol:
1. Reference Standard:
   - The standard answer is definitive and always correct
   - The question is perfectly valid - never question them
   - Do not regenerate answers; only compare with the given standard

2. Comparison Method:
   - Carefully analyze the question's requirements and the standard answer's structure
     * Determine whether the question expects exact matching of the entire standard answer or allows partial matching of its components.
     * This determination must be made based on the question's phrasing and the nature of the standard answer.
   - Compare ONLY the candidate's final answer (ignore all reasoning/explanation errors)
   - Disregard any differences in formatting or presentation style
   - For mathematical expressions: calculate step by step whether the two formulas are equivalent
   - For multiple-choice questions: compare only the final choice and corresponding option content

3. Multi-part Answers:
   - For questions requiring multiple responses (e.g., multi-select):
   - All parts must match the standard answer exactly. 
   - Compare each sub-answer step by step. Partial matches are considered incorrect.

4. Validity Check:
   - Reject answers that are:
     * Incomplete (cut off mid-sentence in the final sentence, lacking a complete response) - Label as INCOMPLETE
     * Repetitive (repetition of words or phrases in a loop) - Label as REPETITIVE
     * Explicit refusals (e.g., directly return "I cannot answer/provide/access ...") - Label as REFUSAL
   - For invalid answers, specify the type in the judgment (e.g., \boxed{C} - INCOMPLETE).

Grading Scale:
\boxed{A} - CORRECT: 
   - Answer matches standard exactly (including equivalent expressions)
   - For numerical answers: consider as equivalent if values match when rounded appropriately
   - Semantically equivalent responses

\boxed{B} - INCORRECT:
   - Any deviation from standard answer
   - Partial matches for multi-part questions

\boxed{C} - INCOMPLETE/REPETITIVE/REFUSAL:
   - Fails validity criteria above (must specify: INCOMPLETE/REPETITIVE/REFUSAL)

Execution Steps and Output Formats:

Analysis step by step: [
Thoroughly evaluate the candidate's answer including:
(1) First check if the answer is INCOMPLETE (cut off mid-sentence), REPETITIVE (looping repetition), or a REFUSAL (explicit denial) - if so, immediately classify as \boxed{C} with the corresponding type.
(2) Analyze the question's core requirements and the standard answer's structure, for example:
- Strict requirements: Identify mandatory constraints (e.g., simplification, answer order, multi-part completeness)
- Tolerant allowances: Ignore non-critical deviations (e.g., missing option labels in MCQs, equivalent but unformatted expressions)
- Required answer type, precision level, etc.
(3) Perform a detailed comparison between the candidate's final answer and the standard answer, for example:
- Content equivalence
- Permitted variations in numerical precision
- Allowed expression formats]
Final Judgment: \boxed{A/B/C} - <CORRECT/INCORRECT/INCOMPLETE/REPETITIVE/REFUSAL>

Here is your task.
<Original Question Begin>
{question}
<Original Question End>

<Standard Answer Begin>
{gold_answer}
<Standard Answer End>

<Candidate's Answer Begin>
{llm_response}
<Candidate's Answer End>

Analysis step by step and Final Judgment:
\end{lstlisting}

\begin{lstlisting}[caption={Prompt B for CoT answer verification}]
As a grading expert, your task is to determine whether the candidate's final answer matches the provided standard answer. Follow these evaluation guidelines precisely:

Evaluation Protocol:
1. Reference Standard:
   - The standard answer is definitive and always correct
   - The question is perfectly valid. Never question them
   - Do not regenerate answers; only compare with the given standard answer

2. Thoroughly evaluate the candidate's answer follow these steps
   - Carefully analyze the question's content and requirements
     * Strict requirements: Identify mandatory constraints (e.g., simplification, answer order, multi-part completeness)
     * Tolerant requirements: Ignore non-critical deviations (e.g., missing option labels in MCQs, equivalent but unformatted expressions)
   - Carefully analyze the standard answer's content and structure. Determine whether the question expects exact matching of the entire standard answer or allows partial matching of its components
   - Validity Check for the candidate's answer. Reject answers that are:
     * Incomplete (cut off mid-sentence in the final sentence, lacking a complete response) - Label as INCOMPLETE
     * Repetitive (repetition of words or phrases in a loop) - Label as REPETITIVE
     * Explicit refusals (e.g., directly return "I cannot answer/provide/access ...") - Label as REFUSAL
   - Perform a detailed comparison between the candidate's final answer and the standard answer
     * Compare ONLY the candidate's final answer (ignore all reasoning/explanation errors)
     * Disregard any differences in formatting or presentation style
     * For mathematical expressions: calculate step by step whether the two formulas are equivalent
     * For multiple-choice questions: compare only the final choice and the corresponding option content
     * For questions requiring multiple sub-answers (e.g., multi-select): All parts must match the standard answer exactly. Compare each sub-answer step by step. Partial matches are considered incorrect.

3. Grading Scale:
   \boxed{A} - CORRECT: 
      - Answer matches standard exactly (including equivalent expressions)
      - For numerical answers: consider as equivalent if values match when rounded appropriately
      - Semantically equivalent responses
   \boxed{B} - INCORRECT:
      - Any deviation from standard answer
      - Partial matches for multi-part questions
   \boxed{C} - INCOMPLETE/REPETITIVE/REFUSAL:
      - Fails validity criteria above (must specify: INCOMPLETE/REPETITIVE/REFUSAL)

Output Formats:
Analysis: [Analysis and evaluate step by step here.]
Final Judgment: \boxed{A/B/C} - <CORRECT/INCORRECT/INCOMPLETE/REPETITIVE/REFUSAL>

Here is your task.
<Original Question Begin>
{question}
<Original Question End>

<Standard Answer Begin>
{gold_answer}
<Standard Answer End>

<Candidate's Answer Begin>
{llm_response}
<Candidate's Answer End>

Analysis:
Final Judgment:
\end{lstlisting}

\begin{lstlisting}[caption={Prompt C for CoT answer verification}]
As a grading expert, your task is to determine whether the candidate's final answer matches the provided standard answer. Follow these evaluation guidelines precisely:

Evaluation Protocol:
1. Reference Standard:
   - The standard answer is definitive and always correct
   - The question is perfectly valid - never question them
   - Do not regenerate answers; only compare with the given standard

2. Comparison Method:
   - Extract ONLY the candidate's final answer (ignore all reasoning/explanation errors)
   - If no complete final answer exists (e.g., response is cut off or contains only reasoning) - INVALID
   - Compare this directly with the standard answer
   - Disregard any differences in formatting or presentation style
   - For mathematical expressions: compare semantic equivalence, not syntax
   - For \boxed{} format: ignore the \boxed notation when comparing

3. Multi-part Answers:
   - For questions requiring multiple responses (e.g., multi-select):
   - All parts must match the standard answer exactly
   - Partial matches are considered incorrect

4. Validity Check:
   - Reject answers that are:
     * Incomplete (cut off mid-response or missing final answer)
     * Purely reasoning without final answer
     * Repetitive or uninterpretable
     * Irrelevant to the question
     * Explicit refusals (e.g., "I cannot answer/provide/access ...")

Grading Scale:
\boxed{A} - CORRECT: 
   - Answer matches standard exactly (including equivalent expressions)
   - For numerical answers: allow 1% tolerance for floating-point variations
   - Semantically equivalent responses

\boxed{B} - INCORRECT:
   - Any deviation from standard answer
   - Partial matches for multi-part questions

\boxed{C} - INVALID:
   - Fails validity criteria above

Execution Steps and Output Formats:
Analysis:
1. Completeness and Validity Check: [confirm if candidate's answer is complete and include the final answer]
2. Extracted Final Answer: [state what was identified as final answer]
3. Standard Comparison: [describe how it matches/mismatches]
Final Judgment: [\boxed{A/B/C}]

Here is your task.
<Original Question Begin>
{question}
<Original Question End>

<Standard Answer Begin>
{gold_answer}
<Standard Answer End>

<Candidate's Answer Begin>
{llm_response}
<Candidate's Answer End>

Analysis and Final Judgment:
\end{lstlisting}

\begin{table}[htbp]
\centering
\caption{List of models used in the experiment with response counts}
\label{tab_list_of_models}
\resizebox{0.7 \textwidth}{!}{
\begin{tabular}{llr}
\toprule
\textbf{Model Family} & \textbf{Model Name} & \textbf{Response Count} \\
\midrule
Yi  & Yi-Lightning & 18496 \\
& Yi-1.5-9B-Chat & 17722 \\
\midrule
GPT  & GPT-4o & 18495 \\
& GPT-4o-mini & 44502 \\
& GPT-4-1-2025-0414 & 2673 \\
& GPT-4.5-preview-2025-02-27 & 18381 \\
\midrule
Doubao & Doubao-Pro-32k-241215 & 6378 \\
& Doubao-Pro-1.5-32k-250115 & 18517 \\
& Doubao-Pro-32k-240828 & 5692 \\
\midrule
Qwen & Qwen-Max-0919 & 18434 \\
& Qwen-Max-2025-01-25 & 29173 \\
& Qwen2.5-Max & 18320 \\
& Qwen2.5-7B-Instruct & 49003 \\
& Qwen2.5-14B-Instruct & 32116 \\
& Qwen2.5-32B-Instruct & 37477 \\
& Qwen2.5-72B-Instruct & 37568 \\
& QwQ-32B & 20623 \\
\midrule
Gemini  & Gemini-2.0-Flash-Exp & 17303 \\
& Gemini-1.5-Pro & 18429 \\
& Gemini-2-5-Pro-03-25 & 669 \\
\midrule
DeepSeek-R1 & DeepSeek-Chat-R1 & 16556 \\
& DeepSeek-R1-distill-Qwen-1.5B & 16012 \\
& DeepSeek-R1-distill-Qwen-7B & 16364 \\
& DeepSeek-R1-distill-Llama-8B & 15731 \\
& DeepSeek-R1-distill-Qwen-14B & 16671 \\
& DeepSeek-R1-distill-Qwen-32B & 16042 \\
& DeepSeek-R1-distill-Llama-70B & 15772 \\
\midrule
Llama & Llama-3-1-8B-Instruct & 44857 \\
& Llama-3-1-70B-Instruct & 18018 \\
& Llama-3-2-3B-Instruct & 28618 \\
& Llama-3-3-70B-Instruct & 28307 \\
\midrule
Mixtral  & Mistral-Small-Instruct-2409 & 18233 \\
& Mistral-Small-3.1-24B-Instruct & 14331 \\
& Ministral-8B-Instruct-2410 & 17962 \\
& Mixtral-Large-Instruct-2411 & 18381 \\
\midrule
Claude  & Claude-3-5-Sonnet-20241022 & 18521 \\
& Claude-3-7-Sonnet-20250219 & 18474 \\
& Claude-3-7-Sonnet-20250219-Thinking & 4723 \\
\midrule
Gemma  & Gemma-2-9B-It & 34541 \\
& Gemma-2-27B-It & 34704 \\
& Gemma3-27B-It & 13120 \\
\midrule
DeepSeek-Chat & DeepSeek-V2.5 & 31896 \\
& DeepSeek-Chat-V3 & 31950 \\
\midrule
InternLM  & InternLM2.5-7B-Chat & 43336 \\
& InternLM2.5-20B-Chat & 37594 \\
& InternLM3-8B-Instruct & 15976 \\
\midrule
Phi & Phi-4 & 18360 \\
\midrule
GLM  & GLM-4-9B-Chat & 17537 \\
& GLM-4-Plus & 18486 \\
\midrule
MiniMax  & MiniMax-Text-01 & 39570 \\
\midrule
Moonshot  & Moonshot-V1-32k & 18067 \\
\midrule
Hunyuan  & Hunyuan-Standard-256K & 18082 \\
\midrule
StepFun & Step-2-16k & 18405 \\
\bottomrule
\end{tabular}}
\end{table}

\clearpage
\subsection{Details of Training Data for CompassVerifier}
\label{Details of CompassVerifier Model Train Data}
For the composition of CompassVerifier train dataset, we use 54420 consist samples from the VerifierBench pipeline as shown in \Cref{fig_pipeline} as the base train set, we then use \textbf{Error-Driven Adversarial Augmentation} and \textbf{Complex Formula Augmentation} to construct extra data comprehensively enhance the capabilities of the CompassVerifier model. The composition of our train data list in \Cref{tab:training_data_composition}.
\begin{table}[h!]
    \centering
    \caption{Composition of CompassVerifier training data}
    \label{tab:training_data_composition}
    \begin{tabular}{lrr}
        \toprule
        Data Source & Number of Samples & Percentage (\%) \\
        \midrule
        Base Train Set (VerifierBench) & 54,420 & 56.20 \\
        Error-Driven Adversarial Augmentation & 24,294 & 25.09 \\
        Complex Formula Augmentation & 18,118 & 18.71 \\
        \midrule
        \textbf{Total} & \textbf{96,832} & \textbf{100.00} \\
        \bottomrule
    \end{tabular}
\end{table}
\paragraph{Error-Driven Adversarial Augmentation.}
Using DeepSeek-v3, we generate 34 Meta-Judge Templates covering common and extreme error scenarios then generate 224294 synthetic examples that emphasize decision boundary cases, especially where human judges tolerate minor errors that baseline verifiers over-penalize. 

\paragraph{Complex Formula Augmentation.}
Applying this augmentation pipeline, we have synthesized approximately 18118 enhanced examples spanning 14 distinct scientific and engineering disciplines.



\subsection{Details of CompassVerifier-as-Reward Experimental Settings} \label{app:details-verifier-as-reward}

\paragraph{Base LLMs.}
We utilize Qwen3-4B-Base \citep{yang2025qwen3} as the base LLM for the GRPO training.

\paragraph{Training Template.} 
We utilize the following training template to prompt the base LLM to generate a response for each question. We only verify the format correctness to ensure the final answer is encapsulated within `$\backslash$boxed\{...final answer...\}'.

\begin{promptbox}{Training Template of CompassVerifier}
    A conversation between a User and an Assistant. The User poses a question, and the Assistant provides a solution. The Assistant's response follows these structured steps:

    1. **Reasoning Process**: The Assistant comprehensively thinks about the problem through a reasoning process.
    
    2. **Conclusion**: The Assistant reaches a conclusion, which is enclosed within `<conclusion>' and `</conclusion>' tags. The final answer is highlighted within `$\backslash$boxed\{...final answer...\}'.
    
    3. **Response Format**: The complete response should be formatted as:
    
    ...reasoning process...
    
    <conclusion>
    
    ...conclusion...
    
    The answer is $\backslash$boxed\{...final answer...\}
    
    </conclusion>
\end{promptbox}

\paragraph{Training Data.}
We utilize the challenging mathematical reasoning dataset Open-S1 \citep{dang2025reinforcementlearningreasoningsmall} as the RL training corpus.
To increase the difficulty of our validation, we curate the final training set by specifically excluding problems with integer solutions from the original Open-S1 dataset.

\paragraph{Evaluation. }
We employ Math-Verify~\citep{math-verify} as our evaluation tool since the answers in these three benchmarks are readily verifiable, making them particularly well-suited for Math-Verify's verification mechanism.

\paragraph{Reward Design. }
We design a simple reward scheme: 0 for answer errors, and 1 for correct responses.

\paragraph{Training Parameters. }
We utilize the following loss function, with Table \ref{tab:rl-training-parameters} detailing the training parameters:
\begin{equation}
    \begin{aligned}
        \mathcal{L} &= \mathbb{E}_{(q,a) \sim \mathcal{D},\{o_i\}_{i=1}^G \sim \pi_{\theta_{\text{old}}}(\cdot|q)} \\
        & \left[ \frac{1}{\sum_{i=1}^G \vert o_i \vert} \sum_{i=1}^G \sum_{t=1}^{\vert o_i \vert} \min \left( \frac{\pi_\theta \left( o_{i,t} | q, o_{i,<t} \right)}{\pi_{\theta_{\text{old}}}\left( o_{i,t} | q, o_{i,<t} \right)} a_{i,t}, \text{clip} \left( \frac{\pi_\theta \left( o_{i,t} | q, o_{i,<t} \right)}{\pi_{\theta_{\text{old}}}\left( o_{i,t} | q, o_{i,<t} \right)}, 1 - \epsilon_{\text{min}, }, 1 - \epsilon_{\text{max}}\right) a_{i,t} \right) \right],
    \end{aligned}
\end{equation}
where $\mathcal{D}$ denotes the training data, $(q,a)$ represents the question-answer pair, $G$ signifies the group size, and
\begin{equation}
    a_{i,t} = r_i - \text{mean}(\{r_i\}_{i=1}^G).
\end{equation}
In this context, $a_{i,t}$ signifies the advantage of response $o_i$ at the $t$-th position, and $r_i$ denotes the reward of response $o_i$. Essentially, the KL penalty of the original GRPO loss is omitted, and zero mean normalization is employed to estimate the advantage.

\begin{table}[h]
    \centering
    \caption{Training parameters of CompassVerifier as reward experiments.} \label{tab:rl-training-parameters}
    \begin{tabular}{lc}
        \toprule
        \textbf{Parameters} & \textbf{Value} \\
        \midrule
         train batch size & $256$ \\
         train epochs & $2$ \\
         learning rate & $1e\text{-}6$ \\
         max prompt length & $4096$ \\
         max response length & $12288$ \\
         $G$ & $8$ \\
         $\epsilon_{\text{min}}$ & $0.2$ \\
         $\epsilon_{\text{max}}$ & $0.28$ \\
        \bottomrule
    \end{tabular}
\end{table}

\paragraph{Hardware. }
All experiments are conducted on clusters equipped with 8 NVIDIA A800-SXM4-80GB GPUs and Intel(R) Xeon(R) Platinum 8336C CPUs, implementing with veRL~\citep{ShengZYWZZPL025}.

\end{document}